\documentclass[10pt,twocolumn,letterpaper]{article}

\usepackage{iccv}
\usepackage{times}
\usepackage{epsfig}
\usepackage{graphicx}
\usepackage{cuted}
\usepackage{capt-of}
\usepackage{xcolor}
\usepackage{amsmath,amssymb,amsfonts,dsfont,pifont,bm,bbm,mathrsfs,mathtools,nicefrac}
\usepackage{algorithm,algpseudocode,listings}
\usepackage{booktabs,multirow,adjustbox,diagbox,threeparttable}
\definecolor{citeblue}{RGB}{48,111,186}
\usepackage[pagebackref=true,breaklinks=true,colorlinks,citecolor=citeblue,bookmarks=false]{hyperref}
\usepackage[capitalize]{cleveref}
\usepackage[accsupp]{axessibility}

\crefname{section}{Sec.}{Secs.}
\Crefname{section}{Section}{Sections}
\crefname{table}{Tab.}{Tabs.}
\Crefname{table}{Table}{Tables}
\crefname{figure}{Fig.}{Figs.}
\Crefname{figure}{Figure}{Figures}
\crefname{equation}{Eq.}{Eqs.}
\Crefname{equation}{Equation}{Equations}
\newcommand{\z}{{\bm z}}
\newcommand{\x}{{\bm x}}

\newcommand{\w}{{\bm w}}
\newcommand{\ww}{{\mathcal{W}}}
\newcommand{\p}{{\bm p}}
\newcommand{\Loss}{\mathcal{L}}
\newcommand{\E}{\mathbb{E}}

\newcommand{\fx}{\tilde{\x}}

\newcommand{\Prec}{Prec.$\uparrow$}
\newcommand{\Recall}{Recall$\uparrow$}
 \newcommand{\squad}{\hspace{0.5em}} 

\newcommand{\method}{LinkGAN\xspace}

\iccvfinalcopy

\def\iccvPaperID{5093}

\newcommand\nonumfootnote[1]{%
\begingroup%
    \renewcommand\thefootnote{}\footnote{\hspace{-3.7pt}#1}%
    \addtocounter{footnote}{-1}%
\endgroup%
}

\ificcvfinal\fi

\begin{document}

\title{LinkGAN: Linking GAN Latents to Pixels for Controllable Image Synthesis}

\author{Jiapeng Zhu\textsuperscript{$\dagger$*1} \squad Ceyuan Yang\textsuperscript{$\dagger$2} \squad Yujun Shen\textsuperscript{$\dagger$3} \squad Zifan Shi\textsuperscript{*1} \squad Bo Dai\textsuperscript{2} \squad Deli Zhao\textsuperscript{4} \squad Qifeng Chen\textsuperscript{1} \\
  \textsuperscript{1}HKUST
  ~\textsuperscript{2}Shanghai AI Laboratory
  ~\textsuperscript{3}Ant Group 
  ~\textsuperscript{4}Alibaba Group  \\
}

\twocolumn[{
\renewcommand\twocolumn[1][]{#1}
\maketitle
\begin{center}
    \vspace{-15pt}
    \includegraphics[width=0.95\linewidth]{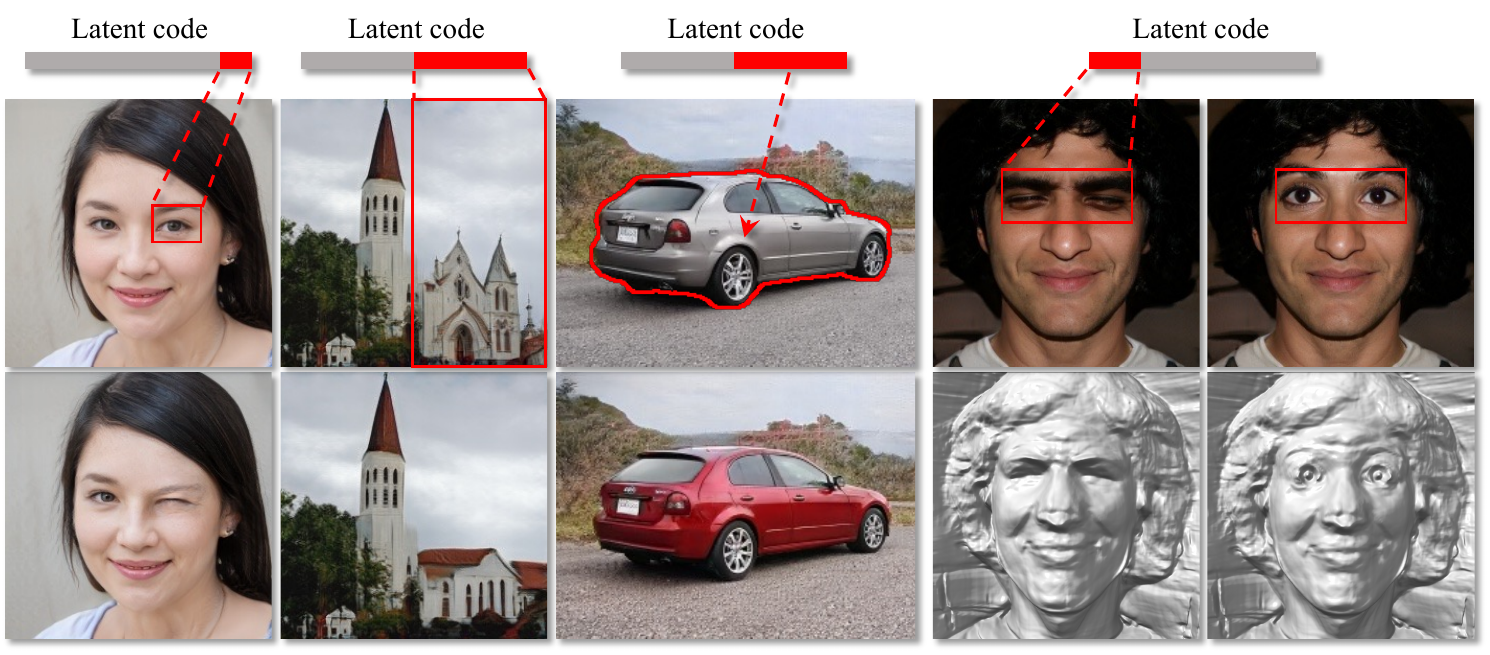}
    \vspace{-2pt}
    \captionof{figure}{%
        \textbf{Precise local control} achieved by \method, where we can manipulate the image content within a spatial region (\textit{e.g.}, a single eye or the right half of the image) or a semantic category (\textit{e.g.}, car) simply by \textit{resampling the latent code on some sparse axes}.
        Our approach works well for 2D image syntheses, like StyleGAN2~\cite{stylegan2} (left three columns), and 3D-aware image synthesis, like EG3D~\cite{Chan2022eg3d} (right two columns).
        It is noteworthy that, under the 3D-aware case, we can control both the appearance and the underlying geometry.
    }
    \label{fig:teaser}
    \vspace{5pt}
\end{center}
}]

\ificcvfinal\thispagestyle{empty}\fi

\begin{abstract}
\vspace{-10pt}

This work presents an easy-to-use regularizer for GAN training, which helps explicitly link some axes of the latent space to a set of pixels in the synthesized image.
Establishing such a connection facilitates a more convenient local control of GAN generation, where users can alter the image content only within a spatial area simply by partially resampling the latent code.
Experimental results confirm four appealing properties of our regularizer, which we call \method.
(1) The latent-pixel linkage is applicable to either a fixed region (\textit{i.e.}, same for all instances) or a particular semantic category (\textit{i.e.}, varying across instances), like the sky.
(2) Two or multiple regions can be independently linked to different latent axes, which further supports joint control. 
(3) Our regularizer can improve the spatial controllability of both 2D and 3D-aware GAN models, barely sacrificing the synthesis performance.
(4) The models trained with our regularizer are compatible with GAN inversion techniques and maintain editability on real images.
The project page can be found \href{https://zhujiapeng.github.io/linkgan/}{here}.
\nonumfootnote{$\dagger$ indicates equal contribution.\\
\indent {\hspace{1mm}}* This work was done during an internship at Ant Group.}

\end{abstract}

\vspace{-10pt}

\section{Introduction}\label{sec:intro}

Generative adversarial networks (GANs)~\cite{gan} have been shown to produce photo-realistic and highly diverse images, facilitating a wide range of real world applications~\cite{pix2pix2017, park2019SPADE, vqgan, ramesh2021zero, shen2020interfacegan}.
The generator in a GAN is formulated to take a randomly sampled latent code as the input and output an image with a feed forward network.
Given a well-learned GAN model, it is generally accepted that a variety of semantics and visual concepts automatically emerge in the latent space~\cite{yang2019semantic, shen2021closed, ganspace, gansteerability, zhu2022resefa}, which naturally support image manipulation.
Some recent work also reveals the potential of GANs in local editing by steering the latent code along a plausible trajectory in the latent space~\cite{ling2021editgan, zhu2021lowrankgan}.

However, most studies on the relationship between the latent codes and their corresponding images depend on a posterior discovery, which usually suffers from three major drawbacks.
(1) Instability: The identification of emerging latent semantics is very sensitive to the samples used for analysis, such that different samples may lead to different results~\cite{ganspace, shen2020interfacegan}.
(2) Inaccuracy: Given the high-dimensional latent space (\textit{e.g.}, 512$d$ in the popular StyleGAN family~\cite{stylegan, stylegan2}), finding a semantically meaningful subspace can be challenging.
(3) Inflexibility: Existing manipulation models are usually linear (\textit{i.e.}, based on vector arithmetic~\cite{shen2020interfacegan, gansteerability}), limiting the editing diversity.

This work offers a new perspective on learning controllable image synthesis.
Instead of discovering the semantics from pre-trained GAN models, we introduce an efficient regularizer into the training of GANs, which is able to \textit{explicitly} link some latent axes with a set of image pixels.  % an image region or a semantic category (\textit{e.g.}, sky) in the synthesis, as shown in \cref{fig:framework}.
In this way, the selected axes and the remaining axes are related to the in-region pixels and out-region pixels, respectively, with little cross-influence (see \cref{fig:teaser}).
Such a design, termed as \textbf{\method}, enables a more accurate and more convenient control of the generation, where we can alter the image content within the linked region simply by \textit{resampling on the corresponding axes}.

We conduct experiments on various datasets to evaluate the efficacy of \method and demonstrate its four appealing properties.
(1) It is possible to link an arbitrary image region to the latent axes, no matter the region is pre-selected before training and fixed for all instances, or refers to a semantic category and varies across instances  (see \cref{exp:single-region}).
(2) Our regularizer is capable of linking multiple regions to different sets of latent axes independently, and allows joint manipulation of these regions (see \cref{exp:multiple-region}).
(3) Our approach lends itself well to both 2D image synthesis models~\cite{stylegan2} and 3D-aware image synthesis models~\cite{Chan2022eg3d}, appearing as sufficiently improving the controllability yet barely harming the synthesis performance.
(4) The models trained with our regularizer are compatible with GAN inversion techniques~\cite{zhu2016generative} and maintain the editability on real images (see \cref{subsec:applications}).
We believe that this work makes a big step towards the spatial controllability of GANs as well as the explicit disentanglement of GAN latent space.
It can be expected that the \textbf{\textit{new characteristic}} (\textit{i.e.}, the latent-pixel linkage) of generative models could open up more possibilities and inspire more applications in the future.

\section{Related Work}\label{sec:related-work}

\noindent\textbf{Generative adversarial networks (GANs)}
are composited by a generator and a discriminator, which are trained simultaneously by playing a two-player minimax game~\cite{gan}, have made tremendous progress in generating high quality and diverse images~\cite{stylegan, stylegan2, stylegan3, biggan, yang2022improving}.
In turn, there are widely used in a variety of tasks, such as representation learning~\cite{donahue2019bigbigan, xu2021generative}, image-to-image translation~\cite{pix2pix2017, choi2020starganv2}, image segmentation~\cite{zhang2021datasetgan}, 3D generation~\cite{xu2021volumegan, Chan2022eg3d, shi20223daware}, \textit{etc.}

\noindent\textbf{Regularizers for GAN training.}
Many attempts have been made to regularize GANs during training~\cite{wgangp, whichgan, stylegan2, wei2021oroJaR, he2021eigengan, peebles2020hessian, yang2022improving}.
Some of them try to improve the training stability of GANs by regularizing the gradients of the discriminator~\cite{wgangp, whichgan}, the spectral norm of each layer~\cite{miyato2018spectral}, or the singular values of the generator~\cite{odena2018generator}.
Besides, some of them~\cite{peebles2020hessian, wei2021oroJaR, he2021eigengan, stylegan2} aim to improve the disentanglement property of GANs.
For example, \cite{peebles2020hessian, wei2021oroJaR} try to disentangle each component in the latent vectors so that each dimension in the latent codes can only affect one attribute on the output images by adding some regularizers (\textit{e.g.}, Hessian Penalty or Orthogonal Jacobian Regularization).

\noindent\textbf{Image editing with GANs.}
Image editing using GANs includes many different tasks, such as style transfer~\cite{styletransfer, adain}, image-to-image translation~\cite{CycleGAN2017, wang2018pix2pixHD, choi2018stargan, park2019SPADE, park2020swapping}, and semantic image editing using pre-trained GANs~\cite{shen2020interfacegan, yang2019semantic, gansteerability}.
For semantic image editing tasks, one line of work is focused on controlling the image globally~\cite{shen2020interfacegan, yang2019semantic, shen2021closed, ganspace, gansteerability, cherepkov2021navigate, plumerault2020controlling, spingarn2021gan, voynov2020unsupervised, li2020latent, Wang2021Jacobian, latentclr}, and another is focused on controlling the image locally~\cite{suzuki2018spatially, xu2021generative, wu2020stylespace, bau2019gandissection, bau2020rewriting, collins2020editing, kim2021exploiting, kafri2022stylefusion, ling2021editgan, zhu2021lowrankgan, zhu2022resefa}.
For local image control, the straightforward way is to employ region-based feature modification~\cite{suzuki2018spatially, xu2021generative, bau2019gandissection, park2020swapping}, which highly relies on the spatial correspondence between the feature maps and the synthesized images.
An alternative way is to control from the latent space~\cite{collins2020editing, ling2021editgan, zhu2021lowrankgan} yet suffers from limited controllability (\textit{e.g.}, it is hard to close one eye with the other kept open).

\noindent\textbf{Independent latent axis control.}
There are many studies in the literature exploring the independent control of the latent axes of GANs~\cite{donahue2017semantically, gancontrol, hudson2021ganformer, hudson2021ganformer2, giraffe, blockgan, alignlatent, vqgan}, such that we can partially re-configure the generated image through resampling the latent code on some axes.
Among them, \cite{donahue2017semantically, gancontrol} target aligning the latent axes with some image attributes under the supervision of pre-trained attribute classifiers, which treat the entire image as a whole, limiting their applications in local control.
Some attempts have been made towards compositional image synthesis~\cite{hudson2021ganformer, hudson2021ganformer2, giraffe, blockgan}, which employs separate latent codes to take responsibility for the generation of different objects.
LDBR~\cite{hong2020low} introduces block-wise latent space and manages to build a spatial correspondence between per-block latents and image patches.
Infinite image generation~\cite{alignlatent, vqgan}, which is able to expand (\textit{e.g.}, outpainting) the synthesis through sampling the latent code repeatedly, can be viewed as a special type of independent latent axis control.
Compared to previous work, our approach is far more flexible in two folds.
(1) We introduce a simple regularizer into GAN training, with no need for architecture re-designing.
(2) We manage to link the latent axes with an arbitrary set of image pixels.

\section{Method}\label{sec:method}

In this section, we introduce a simple yet effective regularizer such that some latent axes of GANs can be \textit{explicitly} linked to a set of image pixels after training.
We first give a brief introduction of GAN formulation in \cref{subsec:preliminary} and describe how to establish the latent-pixel linkage in \cref{subsec:linkgan}.

\subsection{Preliminaries}\label{subsec:preliminary}

A GAN model consists of a generator $G(\cdot)$ that maps latent vectors $\z \sim p(\z)$ to fake images, $i.e$. $\tilde{\x} = G(\z)$, and a discriminator $D(\cdot)$ that tries to differentiate fake images from real ones.
They are trained in an adversarial manner in the sense that the generator tries to fool the discriminator.  % while the discriminator learns to measure the realness of generated images.
The training loss can be formulated as follows:
\begin{align} 
   \Loss_G & = \E_{p(\tilde{\x})}[f(1 - D(\tilde{\x}))], \label{eq:gan-g} \\
   \Loss_D & =  \E_{p(\x)}[f(D(\x))] - \E_{p(\tilde{\x})}[f(1 - D(\tilde{\x}))], \label{eq:gan-d}
\end{align}
where $p(\x)$ and $p(\tilde{\x})$ are the distributions of real images and synthesized images, respectively.
Here, $f(\cdot)$ is a model-specific function that varies between different GANs.

\subsection{Linking Latents to Pixels}\label{subsec:linkgan}
With the rapid development of manipulation technique, several works~\cite{ling2021editgan, zhu2021lowrankgan} have shown that some subspaces of the latent space (\textit{i.e.}, the $\ww$ space in StyleGAN~\cite{stylegan}) can control local semantics over output images.
Specifically, traversing a latent code within those subspaces results in a local modification in the synthesis.
However, there lacks an explicit connection between the local regions and the specified axes of latent spaces. 
To this end, we propose a new regularizer explicitly linking the axes to arbitrary partitions of synthesized images.

\begin{figure}[t]
  \centering
  \includegraphics[width=1.0\linewidth]{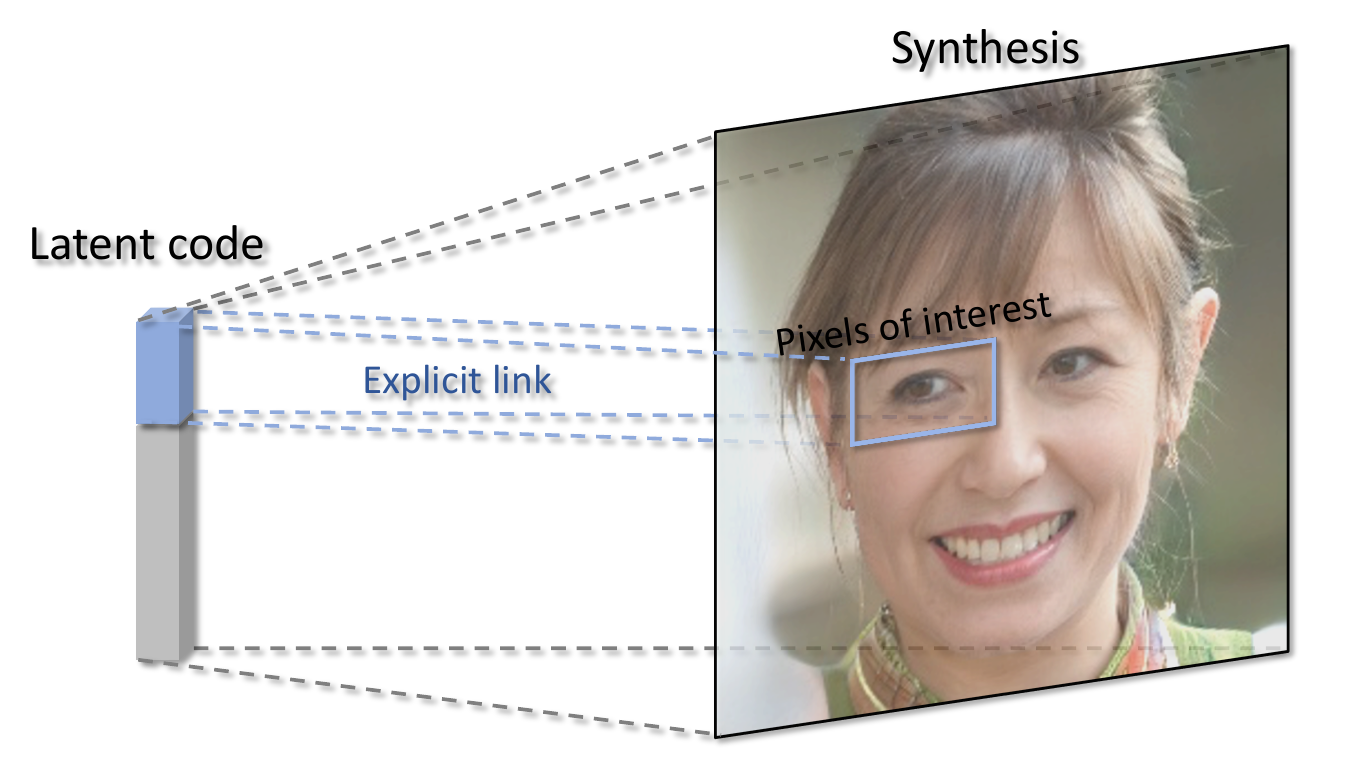}
  \vspace{-10pt}
  \caption{%
    \textbf{Concept diagram} of \method, where some axes of the latent space are \textit{explicitly} linked to the image pixels of a spatial area.
    In this way, we can alter the image content within the linked region simply by resampling the latent code on these axes.
  }
  \label{fig:framework}
  \vspace{-5pt}
\end{figure}

\vspace{5pt}
\noindent \textbf{Partition of latent codes and images.}
In order to set up the explicit link between some axes of latent space and local regions of an image, we first introduce some notations for the corresponding partition. 
Taking StyleGAN~\cite{stylegan} as an example, $\w \in \mathbb{R}^{d_w}$ is the intermediate latent vector of dimension $d_w$ derived from the mapping network.
Through a generator $G(\cdot)$, an image $\fx \in \mathbb{R}^{H \times W \times C}$ is produced, \textit{i.e.}, $\fx = G(\w)$, and we denote $d_x = H \times W \times C$ as the dimension of  $\fx$.
We first divide the latent space into several subspaces.
Namely, a latent code $\w$ could be divided into $K$ partitions and each partition consists of multiple channels, \emph{i.e.}, $\w = [\w_1, \w_2, \dots, \w_K]$, where $ \w_i \in \mathbb{R}^{n_i} $ and $\Sigma_{i=1}^{K} n_i = d_w$.
Similarly, an image $\fx$ could also produce several partitions \emph{i.e.,} $\fx = [\fx_1, \fx_2, \dots, \fx_K]$, where $\fx_i \in \mathbb{R}^{m_i}$ and $\Sigma_{i=1}^{K} m_i = d_x$.
For convenience, we further define $\w_i^c$ and $\fx_i^c$ are the complements of $\w_i$ and $\fx_i$, respectively, \textit{i.e.}, $\w_i^c = [\w_1, \dots, \w_{i-1}, \w_{i+1}, \dots, \w_K]$, $\fx_i^c = [\fx_1, \dots, \fx_{i-1}, \fx_{i+1}, \dots, \fx_K]$.
\cref{fig:framework} presents an example ($K$ is equal to 2) where the blue part of the latent code and pixels within the blue bounding box denote the partitions $\w_i$ and $\fx_i$, respectively. 
Now, our goal is that the latent fragment $\w_i$ only controls the pixels in $\fx_i$ and $\w_i^c$ controls the pixels in $\fx_i^c$, namely, building an explicit link.

%%%% Framework

\vspace{5pt}
\noindent\textbf{Learning objectives.}
To our surprise, we find in practice that a simple regularizer combined with the StyleGAN framework is sufficient to achieve this goal. 
Formally, we can randomly perturb $\w_i$ and $\w_i^c$ and then minimize the variations on $\fx_i^c$ and $\fx_i$, respectively, expecting that $\w_i$ merely controls $\fx_i$ and hardly affects $\fx_i^c$ and vice versa.
Specifically, we can perturb the $\w_i$ partition among $\w$ by given vector $\p_i$ sampled from a standard Gaussian distribution $\mathcal{N}(\bm 0, \bm I^{n_i})$ and get the perturbed image, \textit{i.e.}, $\fx_{1} = G'_{\w}(\w_i, \alpha \p_i) \triangleq G([\w_1, \dots, \w_{i-1}, \w_i + \alpha \p_i, \w_{i+1}, \dots, \w_{K}])$, where $\alpha$ is the perturbation strength.
Furthermore, we can get the perturbed image using a vector $\p_i^{c} \in \mathcal{N}(\bm 0, \bm I^{d_w -n_i})$ to perturb $\w_i^{c}$, \textit{i.e.}, $\fx_{2} = G'_{\w}(\w^c_i, \alpha \p^{c}_i)$.
After obtaining the perturbed images, we can compute the variations in each part.
The pixel change in $\fx_i$ after the perturbation by $\p^{c}_i$ can be computed as
\begin{equation} \label{eq:norm-of-bg}
\begin{aligned}
    \Loss_i & = ||M_i \odot (\fx_2 - \fx)||_2^2 \\
            & = ||M_i \odot (G'_{\w}(\w^c_i, \alpha \p^{c}_i)  - G(\w))||_2^2,
\end{aligned}
\end{equation}
where $M_i$ is the binary mask indicating the chosen pixels of interest (\textit{i.e.}, selecting the pixels in the blue box in \cref{fig:framework}), $||\cdot||_2$ denotes the $\ell_2$ norm.
We enforce the pixels to change in the region $\fx_i$ as minimally as possible after the perturbation by $\p^{c}_i$. 
Similarly, the pixels change in $\fx_i^{c}$ after the perturbation by $\p_i$ can be written as 
\begin{equation} \label{eq:norm-of-fg}
\begin{aligned}
     \Loss_i^{c} & = ||M_i^{c} \odot (\fx_1 - \fx)||_2^2 \\ 
                 & = ||M_i^{c} \odot (G'_{\w}(\w_i, \alpha \p_i) - G(\w))||_2^2,
\end{aligned}
\end{equation}
where $M_i^{c}$ is the binary mask denoting the region out of interest. 
These two losses $\Loss_i$ and $\Loss_i^{c}$ can be integrated as a regularizer in the StyleGAN framework
\begin{equation} \label{eq:reg-loss}
    \Loss_{reg}^{i} = \lambda_{1} \Loss_i + \lambda_{2} \Loss_i^{c},
\end{equation}
where $ \lambda_{1}$ and $ \lambda_{2}$ are the weights to balance these two terms.
Therefore, the total loss to train the generator in StyleGAN  can be formulated as 
\begin{equation} \label{eq:g-loss}
    \Loss = \Loss_G + \sum_{i=1}^{k}\Loss_{reg}^{i},
\end{equation}
where $k$ ($1\leq k \leq K$) is the number of links we want to build.
Practically, we could apply the new regularization in a lazy way, in the sense that  $\sum_{j=1}^{k}\Loss_{reg}^{j}$ is calculated once every several iterations (8 iterations in this paper),  greatly improving the training efficiency.
Additionally, the perturbed images would be also fed into the discriminator during training.
\section{Experiments}\label{sec:exp}

\subsection{Experimental Setup}

We conduct extensive experiments to evaluate our proposed method.
We mainly conduct our experiment on StyleGAN2~\cite{stylegan2} and EG3D~\cite{Chan2022eg3d} models.
The datasets we use are FFHQ~\cite{stylegan}, AFHQ~\cite{choi2020starganv2}, LSUN-Church, and LSUN-Car~\cite{yu2015lsun}. 
We also use a segmentation model~\cite{zhou2018semantic} to select pixels with the same semantic (e.g., all the pixels in the sky on LSUN-Church), which is often used by previous work~\cite{bau2019seeing, bau2019gandissection, wu2020stylespace}.
The main metrics we use to qualify our method are Fr\'{e}chet Inception Distance (FID)~\cite{fid} and the masked Mean Squared Error (MSE)~\cite{zhu2021lowrankgan}.
The experiments are organized as follows.
First, \cref{subsec:effectiveness} shows the properties of \method, which can relate an arbitrary region in the image to the latent fragment.
Second, \cref{subsec:applications} gives some applications of our method, such as local control on the 3D generative model, real images, and some comparisons with the baselines.% regarding the editing precision.
At last, an ablation study on the size of the link latent subspace is presented in \cref{subsec:ablation}.
For the experiment details and more results, please refer to the \textit{Appendix}.

%%%% Figure-single-region-control
\begin{figure*}[t]
  \centering
  \includegraphics[width=1.0\linewidth]{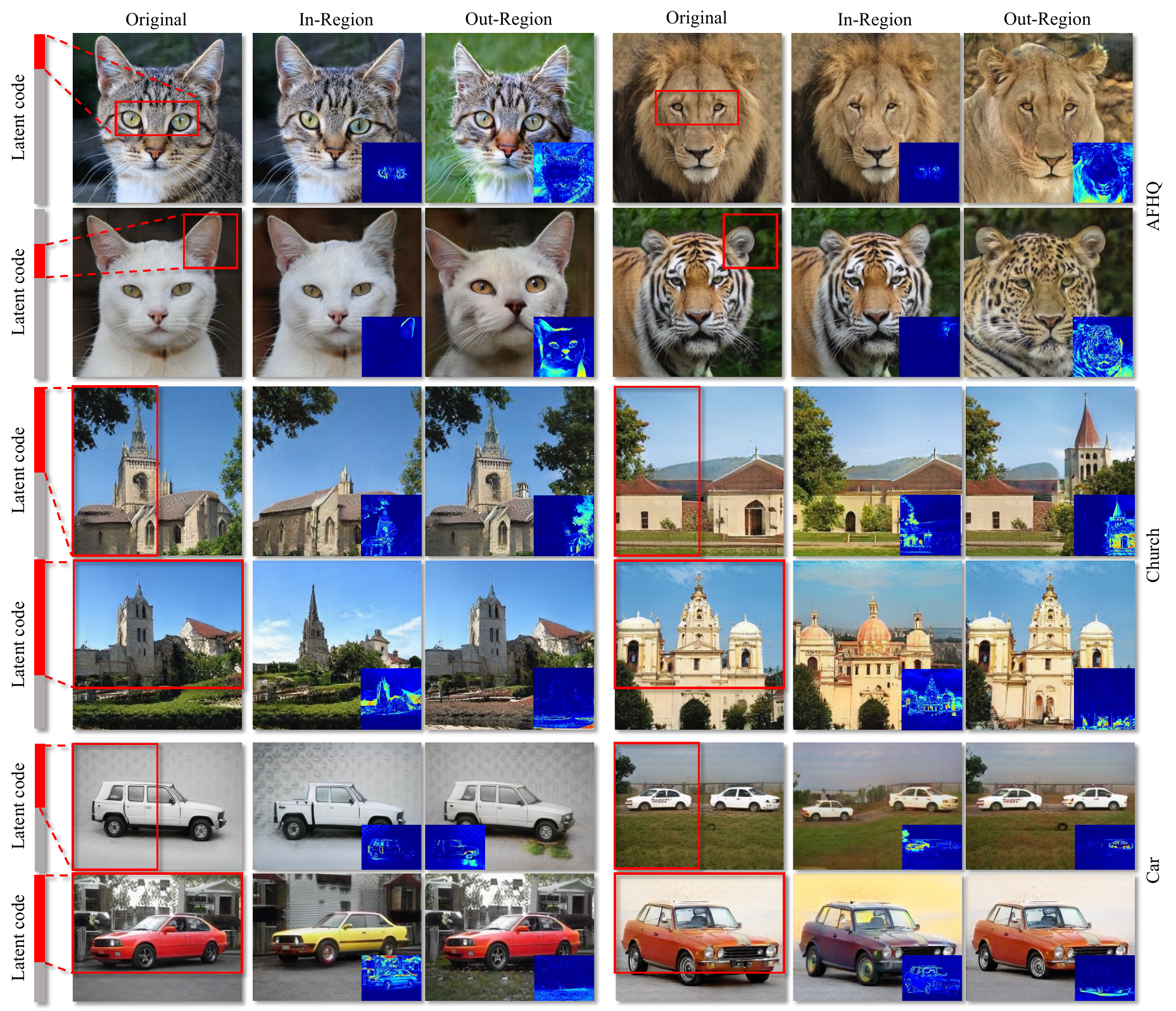}
  \vspace{-10pt}
  \caption{%
    \textbf{Linking latents to single fixed region}, which is pre-selected before training and shared by all instances.
    Linked latent subspaces and regions are highlighted with \textcolor{red}{red} fragments and boxes, respectively, and the heatmaps reflect the change of pixel values after in-region resampling and out-region resampling.
    We find that \method can robustly link the latent to an arbitrary image region. %even semantically meaningless ones (\textit{e.g.}, the second row).
    }
  \label{fig:single-region}
  \vspace{-10pt}
\end{figure*}

%%%% Figure-semantic-region-control
\begin{figure*}[t]
  \centering
  \includegraphics[width=1.0\linewidth]{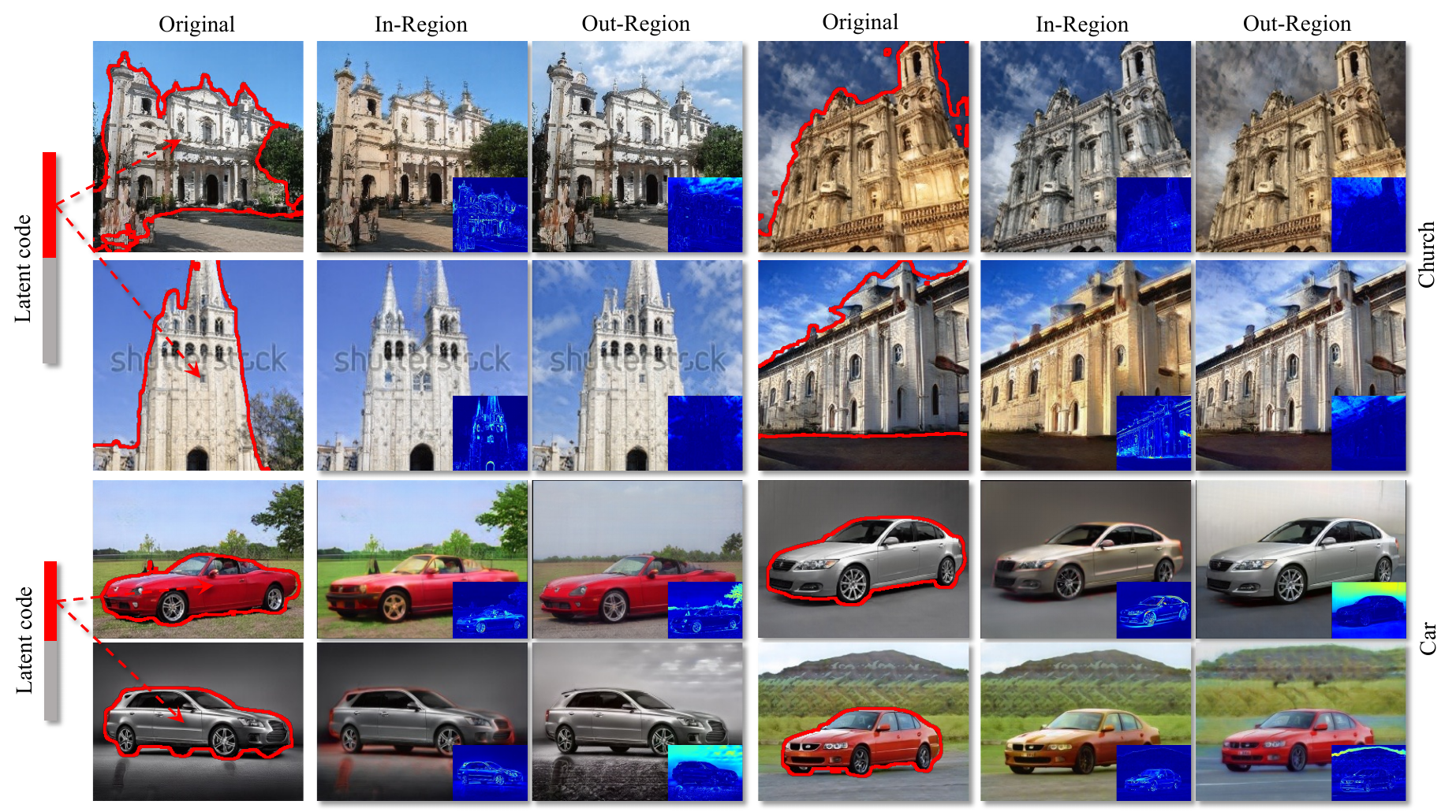}
  \vspace{-10pt}
  \caption{%
    \textbf{Linking latents to the semantic region} (\textit{i.e.}, church and car), which dynamically varies across instances.
    Our \method manages to precisely control a particular semantic category simply by resampling on some sparse latent axes.
    }
  \label{fig:semantic-region}
  \vspace{-10pt}
\end{figure*}

%%%% Figure-multi-regions-control
\begin{figure*}[t]
  \centering
  \includegraphics[width=1.0\linewidth]{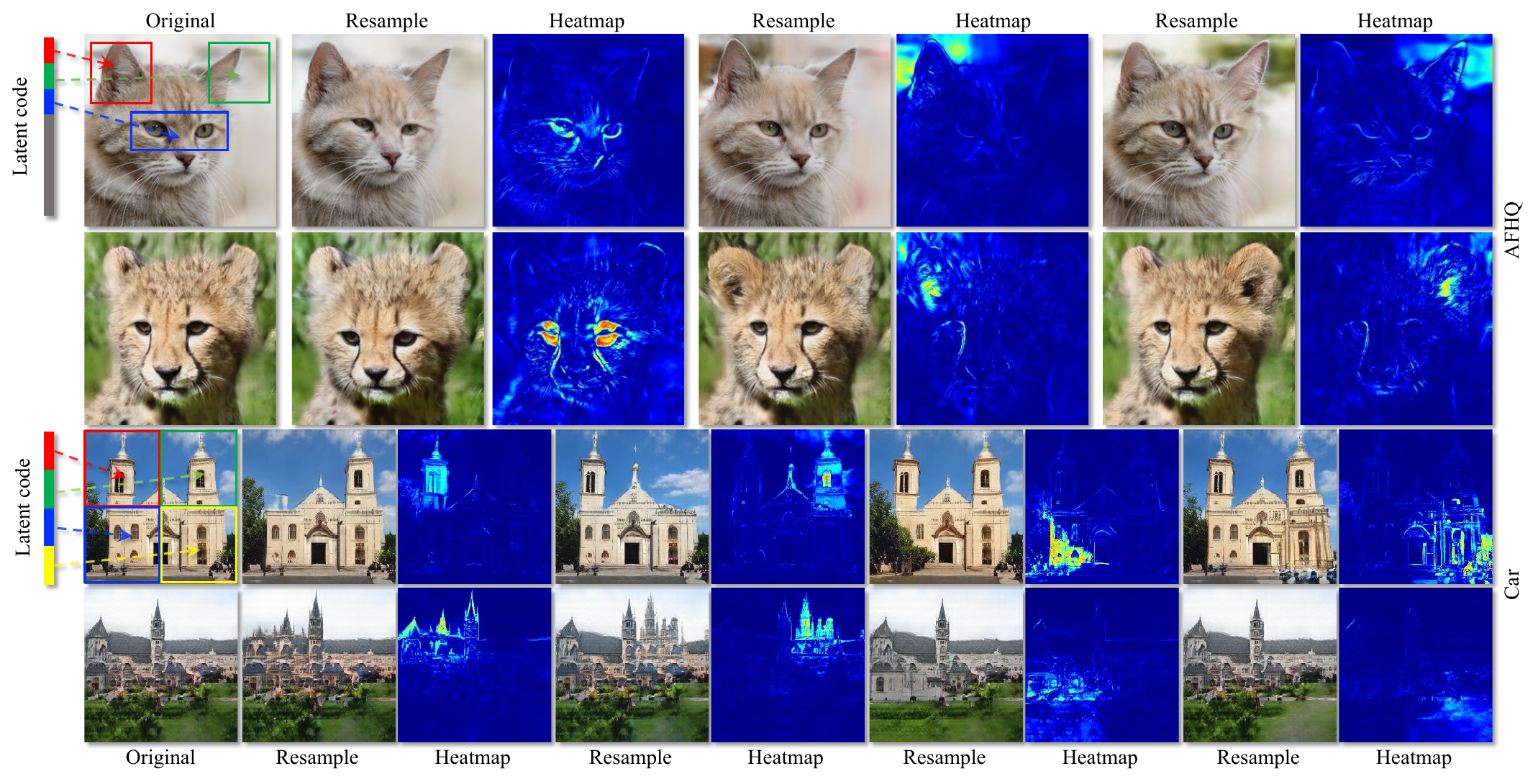}
  \vspace{-10pt}
  \caption{%
    \textbf{Linking latents to multiple regions}, where the linked latent subspaces and image regions are highlighted using different colors.
    Each linked region can be independently controlled by partially resampling the corresponding latent code.
    % simultaneously linked to some non-overlapping sets of latent axes and can be independently controlled by partially resampling the latent codes.
    }
  \label{fig:multi-regions}
  \vspace{-5pt}
\end{figure*}

\subsection{Properties of \method}\label{subsec:effectiveness}

In this section, we mainly demonstrate the effectiveness of the proposed approach by explicitly linking the pixels in any region (both the single region or multi-regions) to a partition of the corresponding latent codes, while seldom deteriorating the quality of synthesis.
\cref{tab:fid} reports FID on different datasets when our regularizer is added, from which we can see our regularizer only has a minor influence on the synthesized quality.
Empirically, we find that it would more stable if the proposed regularizer is incorporated after the convergence of the generator. 
Therefore, we start training from a relatively well-trained generator and equipping it with our approach. 
\vspace{-10pt}

%%%% Table: FID-after-training.
\setlength{\tabcolsep}{6.5pt}
\begin{table}[t]
  \setlength{\tabcolsep}{3.5pt}
  \caption{%
    Performance change after introducing our proposed regularizer into 2D and 3D baselines, where the synthesis quality slightly drops but the controllability significantly improves (see \cref{fig:single-region,fig:semantic-region,fig:multi-regions,fig:eg3d} for details).
  }
  \label{tab:fid}
  \vspace{2pt}
  \centering\small
  \begin{tabular}{l|cccc|c}
    \toprule
              &  \multicolumn{4}{c|}{StyleGAN2~\cite{stylegan2}}
              &  \multicolumn{1}{c}{EG3D~\cite{Chan2022eg3d}} \\
     \hline
     Dataset 
              &  FFHQ  &  AFHQ  &   Car  &  Church  &  FFHQ   \\
    \hline
    LDBR~\cite{hong2020low}   
              &  12.24  &  --  &  --  &  8.68   &  --   \\
              
      \textit{w/o} Linking   
              &  3.98  &  8.44  &  2.95  &   3.82   &  4.28   \\
      LinkGAN (ours)
              &  5.00  &  9.85  &  3.09  &   3.97   &  4.25   \\
    \bottomrule
  \end{tabular}
  \vspace{-8pt}
\end{table}

\subsubsection{Linking Latents to Single Region}\label{exp:single-region}

Regarding the partition of latent codes, we could easily choose the first several channels as one group. Accordingly, the remaining ones become the complementary code. Therefore, the goal of the proposed regularizer is to enable the explicit control of certain regions of interest through the chosen channels. 
Note that the number of first channels that would be grouped usually depends on the area ratio of the chosen region over the entire image. 
In the following context, we will show different ways of choosing pixels out of images and building explicit links between the chosen channels and pixels.

\noindent\textbf{Region-based control.}
One general way of grouping pixels is to use a bounding box that could cover a rectangle region. 
\cref{fig:single-region} presents the qualitative results of choosing different regions randomly. 
Red bounding boxes in \cref{fig:single-region} denote the chosen regions of interest. 
In terms of animal faces on AFHQ, we randomly select two spatial patches and link each region to a specific latent fragment (\emph{e.g.}, the latent fragment can be localized at a random position).
Obviously, after building the explicit link, we could merely change the chosen regions by perturbing the corresponding partition of latent codes, while maintaining the rest regions untouched. 
Besides, perturbing the complementary latent codes results in substantial change for regions out of interest, demonstrating that the spatial controlling is well-built by the proposed explicit link. 
Additionally, we also verify the effectiveness of our regularizer on various datasets. 
For instance, the connection between a partition of latent code and half of the entire image (\textit{i.e.}, Church and Car) also could be easily set up, causing appealing editing results.
The LSUN Church and Car results imply that even if the images are not aligned, we can still build a link and get satisfying editing results. 
In other words, whether images are aligned does not affect the linkage construction.
The difference maps further present how well such an explicit link could control a region of interest.

%%%% Figure: Controlling-on-3D
\begin{figure*}[t]
  \centering
  \includegraphics[width=0.98\linewidth]{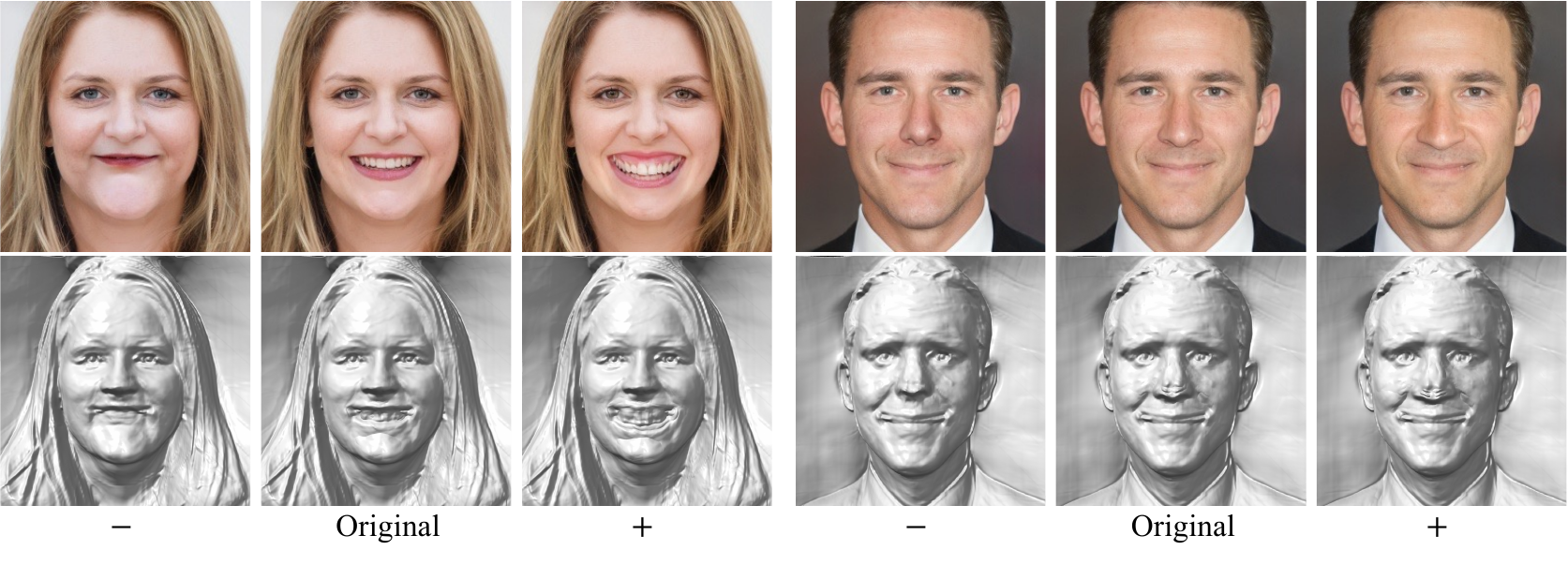}
  \vspace{-5pt}
  \caption{%
    \textbf{Controllability on 3D-aware generative model}, \textit{i.e.}, EG3D~\cite{Chan2022eg3d}, under the cases of mouth and nose.
    We find that \method is well compatible with 3D-aware image synthesis and allows controlling both the appearance and the underlying geometry.
  }
  \label{fig:eg3d}
  \vspace{-5pt}
\end{figure*}

\noindent\textbf{Semantic-based control.}
Prior experimental results demonstrate the control on a rectangle region that seems to be irrelevant to a certain visual concept. Namely, this link is semantic-agnostic since it merely bridges several channels with spatial locations rather than semantics. 
Therefore, we further conduct experiments on semantic controlling. 
To be specific, by leveraging an off-the-shelf segmentation model~\cite{zhou2018semantic}, we could easily obtain mask annotations that specify various semantics.
\cref{fig:semantic-region} presents the semantic control on two datasets, LSUN Church and Car~\cite{yu2015lsun}.
In particular, churches and cars are chosen as the semantics that we would like to build a link between latent space to, no matter where the chosen semantics are. 
Similarly, we manage to connect several channels of latent space with a given semantic such that perturbing the chosen channels will result in the obvious change of semantics. 
For instance, the color and shape of a church vary while the sky keeps the same and vice versa. 
Regarding the experiments on cars, the color could be modified no matter what cars face and how many pixels cars occupy. 
All these results together with the rectangle region control demonstrate the arbitrary region control enabled by our approach.
\vspace{-10pt}

\subsubsection{Linking Latents to Multiple Regions}\label{exp:multiple-region}
After checking the effectiveness of our approach to build one explicit link, a natural question then arises: is it possible to link multiple regions of interest to multiple partitions of latent codes?
The answer is yes.
\cref{fig:multi-regions} presents the corresponding results.
On the top group, we link three subspaces to three image regions \emph{i.e.}, eyes, top-left, and top-right regions, respectively.
Even though we could remain to manipulate semantics individually.
The bottom one moves forward to a more challenging setting where both latent spaces and images are equally divided into four groups and four corners without any overlap.
To this end, we could tell that such a regularizer could build a full explicit link between the entire latent space and the whole synthesis in a disentangled way. 
Namely, we can even tokenize an image and assign one subspace to each token.

\subsection{Applications of \method}\label{subsec:applications}

In this part, we show that our proposed method can be used in various applications, such as controlling 3D generative models, real image manipulation, and precise local image editing, \textit{etc}.

\noindent\textbf{Towards 3D-aware generation.} 
We implement our regularizer on the 3D generative model EG3D~\cite{Chan2022eg3d}.
Surprisingly, our regularizer performs well not only in controlling the RGB images but also in controlling the geometry of the corresponding image, showing the good generalization ability of our regularizer.
\cref{fig:eg3d} shows the results of controlling the mouth and nose region by perturbing the first 64 channels of latent codes.
Importantly, controlling the linked subspace simultaneously changes the RGB images and their geometry, \emph{i.e.}, the mouth is opening for both RGB and corresponding 3D geometry.

\begin{figure}[t]
  \centering
  \includegraphics[width=1.0\linewidth]{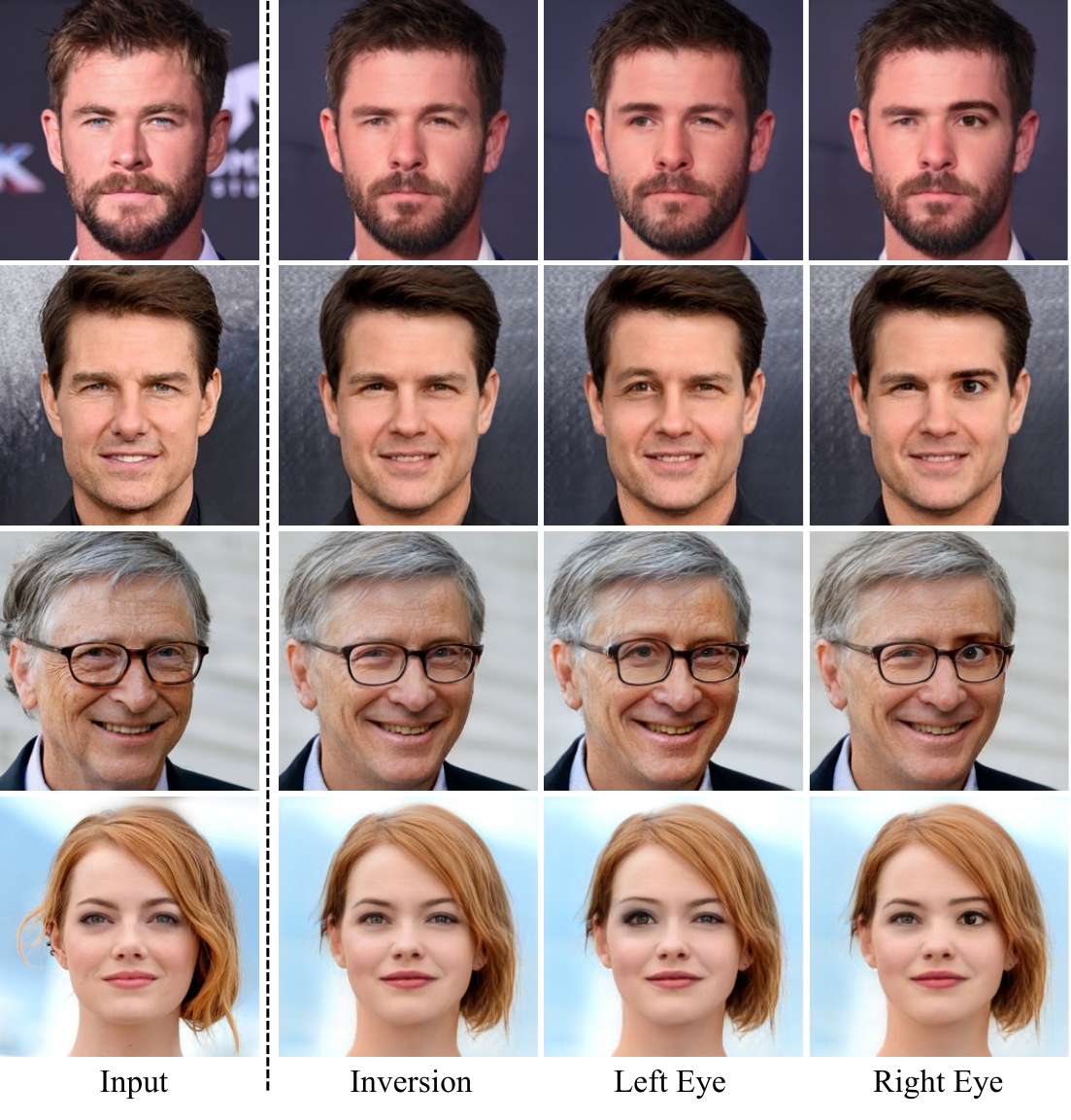}
  \vspace{-10pt}
  \caption{%
    \textbf{Real image editing} achieved by \method via borrowing the GAN inversion technique~\cite{stylegan2}.
    We manage to edit the two eyes of humans independently in a very convenient way, \textit{i.e.}, partially resampling the inverted code.
  }
  \label{fig:real-img}
  \vspace{-10pt}
\end{figure}

\noindent\textbf{Real image editing.}
After the generator is trained, we can use the property of the trained generator to control real images locally by inversion~\cite{zhu2016generative, stylegan2}.
\cref{fig:real-img} shows the editing results on the real image, in which the eyes can be independently controlled, \textit{i.e.}, we can only open one eye yet keep another eye untouched.
In this case, we need to explicitly link two eye regions to two latent subspaces, \textit{i.e.}, one subspace controls one eye.
And when the generator is well-learned, we can edit the eye region by controlling the corresponding subspace on the inverted latent code.

%%%% Figure: Real image editing

%%%% Figure: Compare-with-existing-method
\begin{figure*}[t]
  \centering
  \includegraphics[width=1.0\linewidth]{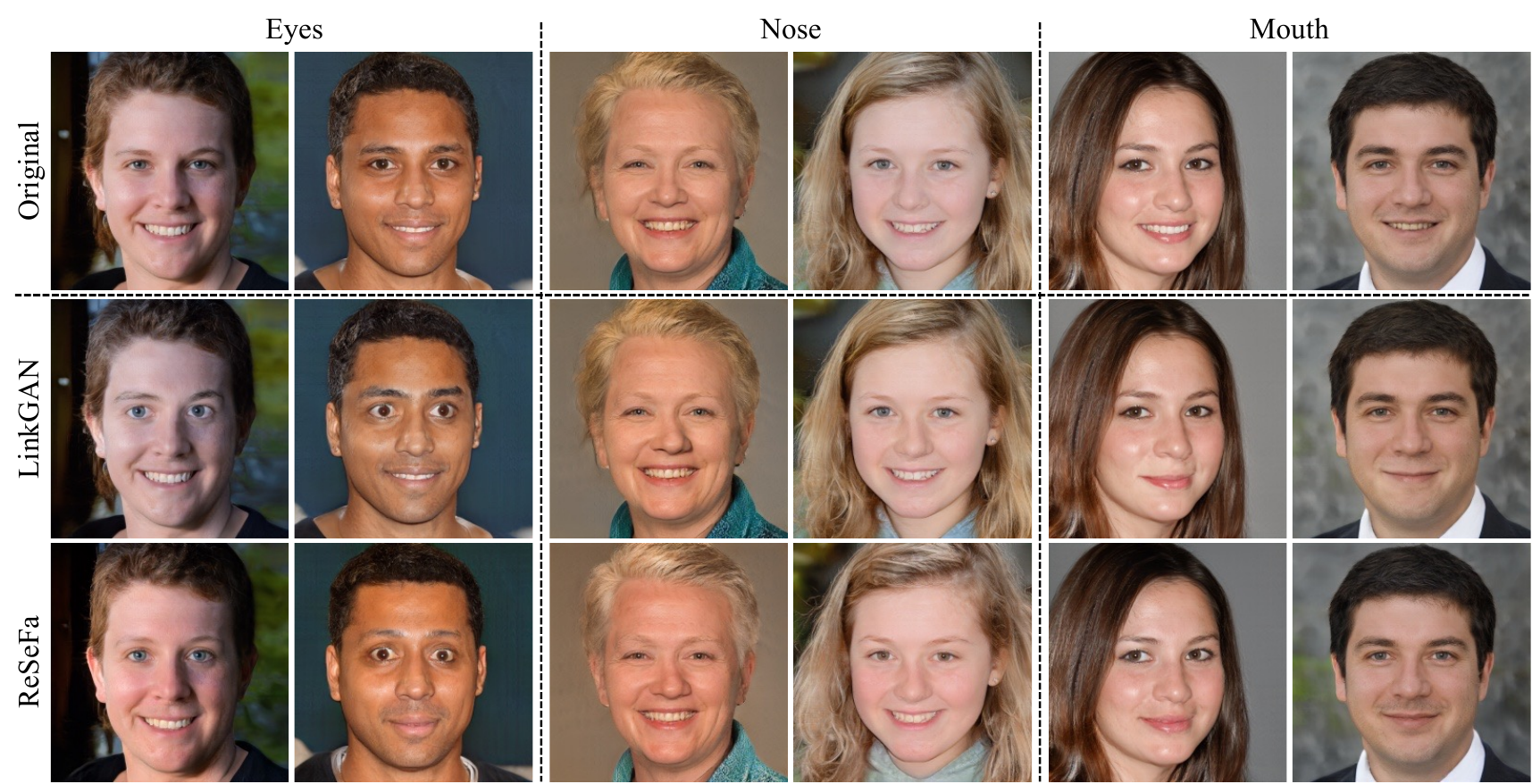}
  \vspace{-10pt}
  \caption{%
    \textbf{Qualitative comparison} with ReSeFa~\cite{zhu2022resefa}, which posteriorly discovers semantics from a pre-trained model, on the task of local editing.
    \method achieves more precise control within the regions of interest.
    See \cref{tab:comparison-baselines} for quantitative results.
  }
  \label{fig:compare}
  \vspace{-10pt}
\end{figure*}

%%%% Table: FID after training.
\begin{table}[t]
  \setlength{\tabcolsep}{2.5pt}
  \caption{%
    \textbf{Quantitative comparison} with baselines on the task of local editing.
    Pixel-wise mean square error (MSE) \textit{within/out of} the region of interest (scaled by $1e^{-3}$ for better readability) is used as the metric.
    Lower $\text{MSE}_{o}$ and higher $\text{MSE}_{i}$ are better.
  }
  \label{tab:comparison-baselines}
  \vspace{5pt}
  \centering\small
  \begin{tabular}{l|cc|cc|cc}
    \toprule
     Region   &  \multicolumn{2}{c|}{Eyes}
              &  \multicolumn{2}{c|}{Nose}
              &  \multicolumn{2}{c}{Mouth} \\
    \hline
    Metrics
              & $\text{MSE}_{i}$ & $\text{MSE}_{o}$ & $\text{MSE}_{i} $   
              & $\text{MSE}_{o}$ & $\text{MSE}_{i}$ & $\text{MSE}_{o} $    \\
    \hline
    StyleCLIP~\cite{styleclip}
              &  3.91  &  74.17  &  1.91  &  72.73  &  3.81 & 65.42   \\
     ReSeFa~\cite{zhu2022resefa}     
              &  5.90  &  61.14  &  1.12  &  60.4  &  2.02  & 50.55   \\
    StyleSpace~\cite{wu2020stylespace}
              &  3.81  &  18.21   &  0.40  &  14.30   &  3.6  & 19.04   \\
     LinkGAN (ours)
              &  5.25  &  2.24    &  1.82  &  2.25   &  3.10  & 2.21    \\
               
    \bottomrule
  \end{tabular}
  \vspace{-5pt}
\end{table}

\noindent\textbf{Comparison with existing methods.}
Now we compare our method with some state-of-the-art algorithms.
We choose LDBR~\cite{hong2020low}, StyleSpace~\cite{wu2020stylespace}, StyleCLIP~\cite{styleclip}, and ReSeFa~\cite{zhu2022resefa} to compare.
For LDBR, we report FID in \cref{tab:fid}, from which we can see that our method significantly outperforms it.%
\footnote{There is no official implementation or released checkpoints. Hence we do not report the qualitative results. The quantitative results on FFHQ and Church are borrowed from the original paper~\cite{hong2020low}.}
And for the rest methods, we compare the accuracy when editing the eyes, nose, and mouth of the face synthesis.
\cref{tab:comparison-baselines} reports the masked MSE between our method and these baselines when controlling those three regions.
Namely, when editing a specific region, we want the change in this region to be as larger as possible (the higher $\text{MSE}_i$, the better) and the change in the remaining region as small as possible (the smaller $\text{MSE}_o$, the better).
For these methods, we can observe that the MSEs within the edited regions are comparable.
However, regarding the MSEs out of the edited regions, our method significantly outperforms these three baselines.
\cref{fig:compare} gives the qualitative comparison with ReSeFa, and for the comparison with other methods, we include them in \textit{Appendix} due to the limited space.
From \cref{fig:compare}, we can observe that our method can reach more precise control on the local regions than ReSeFa.
For instance, when modifying eyes, ReSeFa also results in a change of face color. 
On the contrary, when editing the specific region, our method has negligible changes in the other regions.

\subsection{Ablation Study on Linking Dimensionality}\label{subsec:ablation}
In this part, we conduct an ablation study on how many axes are required to build an explicit link.
Eyes of faces are chosen as regions of interest. 
\cref{tab:abla-mse} gives the quantitative results of changing in/out eye regions with the same perturbation strength.
In \cref{tab:abla-mse}, all the training configurations are the same except for the number of axes during training.
$\text{MSE}_{i}$ and $\text{MSE}_{o}$ are computed in and out of the eye region when perturbing on their complementary latent space, respectively.
Take axes number 8 as an example, the $\text{MSE}_{i}$ is computed within the eye region when perturbing on axes from 8 to 512, while $\text{MSE}_{o}$ is computed out of the eye region perturbing on axes from 0 to 8.
In such a way, precise control could be obtained since the perturbing on the complementary latent space should barely influence the regions of interest.
Hence, in this situation, both $\text{MSE}_{i}$ and $\text{MSE}_{o}$ are the smaller, the better.
Obviously, when occupying the first 64 axes, we can get satisfying results since the sum of them is the smallest.
In practice, we set the number of axes in latent code to 64 in most cases, such as when controlling on eyes, nose, mouth, \textit{etc}.

%%%% Table: MSE in and out region using different axes.
\begin{table}[t]
  \setlength{\tabcolsep}{4.8pt}
  \caption{%
    \textbf{Ablation study} on the linking dimensionality.
    $\text{MSE}_{i}$ measures the effect of unlinked axes on the linked region, while $\text{MSE}_{o}$ measures the effect of linked axes on the unlinked region, both of which enjoy a small value.
    All numbers are scaled by $1e^{-3}$ for better readability.
  }
  \label{tab:abla-mse}
  \vspace{5pt}
  \centering\small
  \begin{tabular}{l|cccccc}
    \toprule
    \# Linked axes 
              &    8    &    16   &   32   &   64   &  128   & 256    \\
             
    \hline
      $\text{MSE}_{i}$   
              &  17.45  &  16.70  &  3.29  &  0.95  &  0.78  & 0.43     \\
      $\text{MSE}_{o}$   
              &  0.86   &  1.53   &  7.41  &  8.20  &  8.71  & 24.78    \\
    \bottomrule
  \end{tabular}
  \vspace{-8pt}
\end{table}

\section{Discussion and Conclusion}\label{sec:conclusion}
We have demonstrated the success of our approach in linkage building, flexible controllability, and more precise spatial control.
Still, there are some limitations.
For example, the built linkage is not perfect, such as when editing a specific part, the remaining area is slightly influenced as the $\text{MSE}_o$ shown in \cref{tab:comparison-baselines}.
The success of this linkage also brings a side effect, \textit{i.e.}, the inconsistency sometimes will appear on the image after we resample part of the latent code, see the detailed analysis in \textit{Appendix}.
In summary, this work proposes \method that explicitly links some latent axes to some specific pixels in the images by utilizing an easy yet powerful regularizer.
Extensive experiments demonstrate the capability of \method in local synthesis control using the precisely linked latent subspace.
{\small
\bibliographystyle{ieee_fullname}
\bibliography{ref}
}

\clearpage
\appendix
\section*{Appendix}

This paper proposes \method that explicitly links some latent axes to a region of an image or a semantic by utilizing an easy yet powerful regularizer.
In this supplementary material, we first give the implementation details of our method in~\cref{sec:implement-details}.
Second, more results are given in~\cref{sec:more-results}, including comparisons with other methods and more quantitative and qualitative results of our methods.
Third, we give an additional ablation study in~\cref{sec:abla} besides the one offered in the main text, \textit{i.e.}, the problem of image inconsistency after resampling.
At last, we give some discussions in~\cref{sec:conclusion-supp}.
\section{Implementation Details}\label{sec:implement-details}
We use the \href{https://github.com/NVlabs/stylegan2-ada-pytorch}{official Pytorch implementation} of StyleGAN2~\cite{stylegan2} and \href{https://github.com/NVlabs/eg3d}{official Pytorch implementation} of EG3D~\cite{Chan2022eg3d} to validate our method.
We keep all the parameters untouched except our newly added regularizer during training. 
We followed the original codebase to compute FID, and for the masked MSE, we calculated it on 10,000 images for each edit.
The update frequency of our lazy regularization is 8.
For how many axes we use to control the specific region, we list below:
1). For small regions, we use 64 axes, such as the eye, nose, mouth, and ear region on FFHQ or AFHQ. 
2). For the larger region, such as the left region of the human face and the bottom part of the church in Fig.3 of the main text, we use 128 axes. 
Also, for linking the whole image of Fig. 5 in the main text along with two (~\cref{fig:res-ffhq}, ~\cref{fig:res-church}) in this appendix, each part has a size of 128 since we evenly split the latent space.
3). When the partition size becomes bigger, such as half of the image, we use 256 axes, and for the semantic control (church, sky, and car) in Fig.4 of the main text, we use 256 axes as well.
For the loss weight $ \lambda_{1}$ and $ \lambda_{2}$, we list as below:
1). For the latent segment with 64 axes, we set $ \lambda_{1}$ equal to 0.04 and $ \lambda_{2}$ equal to 0.01.
2). For the latent segment with 128 axes, we set $ \lambda_{1}$ equal to 0.03 and $ \lambda_{2}$ equal to 0.01.
3). For the latent segment with 256 axes, we set $ \lambda_{1}$ equal to 0.02 and $ \lambda_{2}$ equal to 0.02.
The perturbation strength $\alpha$ is set to one in all the experiments.

\noindent\textbf{Training time.} 
Recall that we are just finetuning the generator from an official checkpoint.
Hence, building such a link will not take much time, which usually takes $ 4 \sim 8 $ hours, depending on the dataset. 
However, achieving a better FID requires a longer training time.
We also do an experiment that trains a GAN from scratch on the FFHQ dataset and then involves the regularizer when the FID is lower than 10, in which we get similar results regarding the controllability and the generation performance compared to finetuning.
In such cases, the train time is roughly equal to the original StyleGAN when getting the smallest FID value for each training.
\section{More results}\label{sec:more-results}

%%%% Figure: Compare-with-existing-method-eyes
\begin{figure*}[t]
  \centering
  \includegraphics[width=1.0\linewidth]{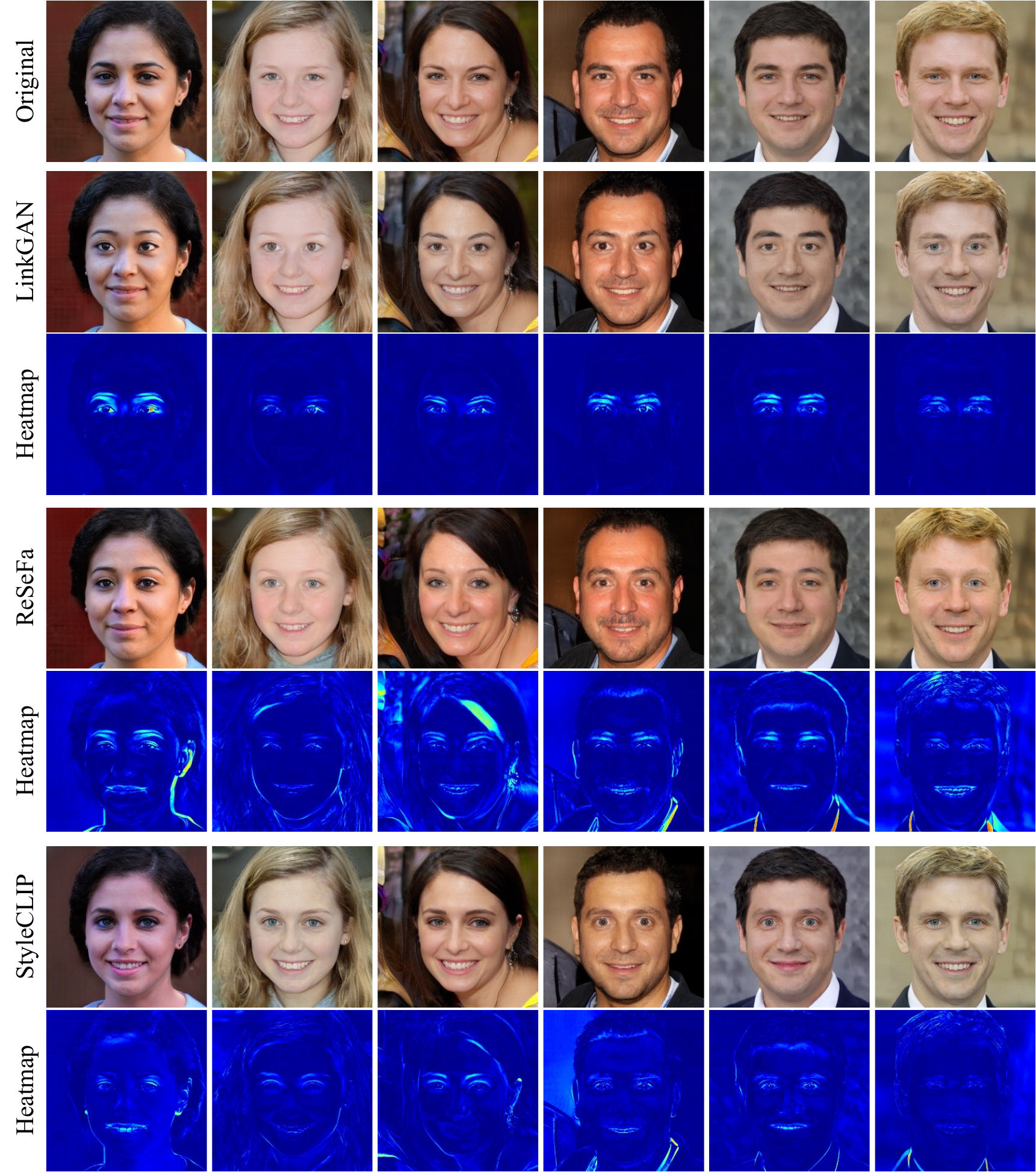}
  \vspace{-5pt}
  \caption{
    \textbf{Qualitative comparison} when manipulating the eyes region with ReSeFa~\cite{zhu2022resefa} and StyleCLIP~\cite{styleclip}.
    As we can see from the heatmaps, \method achieves more precise control within the regions of interest.
  }
  \label{fig:compare-eyes}
  \vspace{-10pt}
\end{figure*}

%%%% Figure: Compare-with-existing-method-eyes
\begin{figure*}[t]
  \centering
  \includegraphics[width=1.0\linewidth]{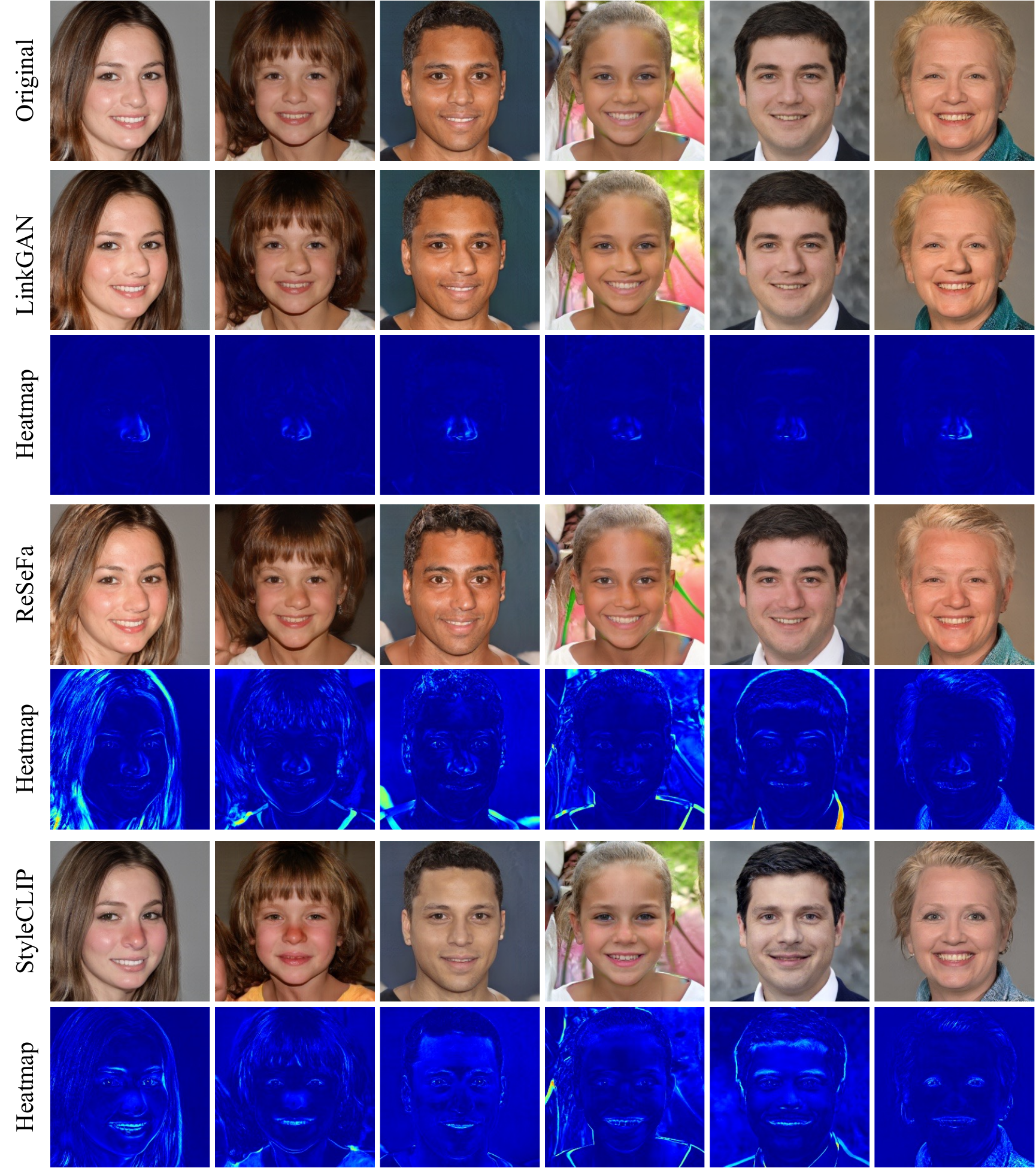}
  \vspace{-5pt}
  \caption{
   \textbf{Qualitative comparison} when manipulating the nose region with ReSeFa~\cite{zhu2022resefa} and StyleCLIP~\cite{styleclip}.
    As we can see from the heatmaps, \method achieves more precise control within the regions of interest.
  }
  \label{fig:compare-nose}
  \vspace{-10pt}
\end{figure*}

%%%% Figure: Compare-with-existing-method-eyes
\begin{figure*}[t]
  \centering
  \includegraphics[width=1.0\linewidth]{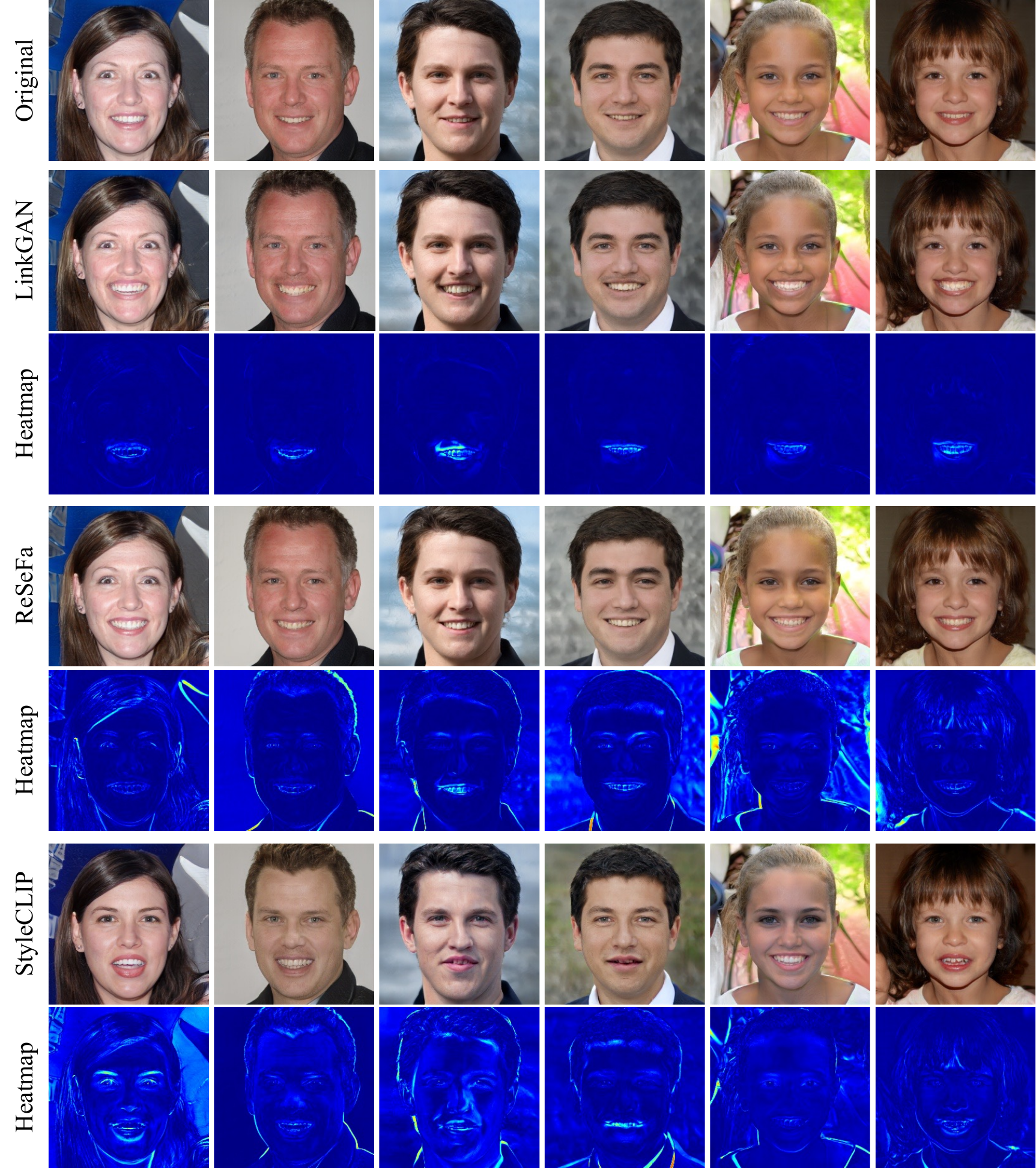}
  \vspace{-5pt}
  \caption{
    \textbf{Qualitative comparison} when manipulating the mouth region with ReSeFa~\cite{zhu2022resefa} and StyleCLIP~\cite{styleclip}.
    As we can see from the heatmaps, \method achieves more precise control within the regions of interest.
  }
  \label{fig:compare-mouth}
  \vspace{-10pt}
\end{figure*}

%%%% Figure: Compare-with-existing-method-eyes
\begin{figure*}[t]
  \centering
  \includegraphics[width=0.83\linewidth]{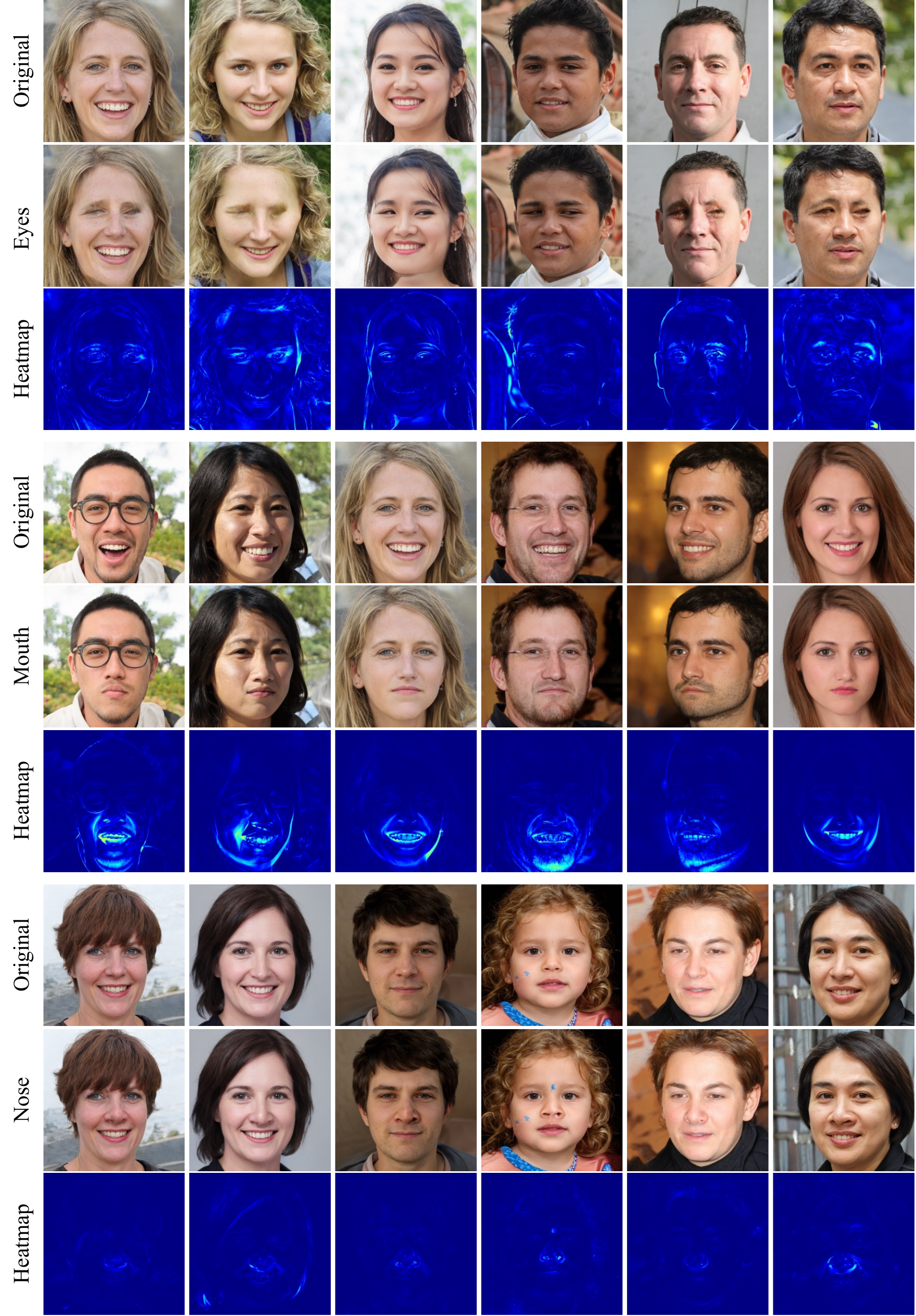}
  % \vspace{-5pt}
  \caption{
    \textbf{Qualitative results} of StyleSpace~\cite{wu2020stylespace} when manipulating on different regions along with corresponding heatmaps.
  }
  \label{fig:compare-stylespace}
  \vspace{-10pt}
\end{figure*}

%%%% Figure: Results on the ffhq.
\begin{figure*}[t]
  \centering
  \includegraphics[width=1.0\linewidth]{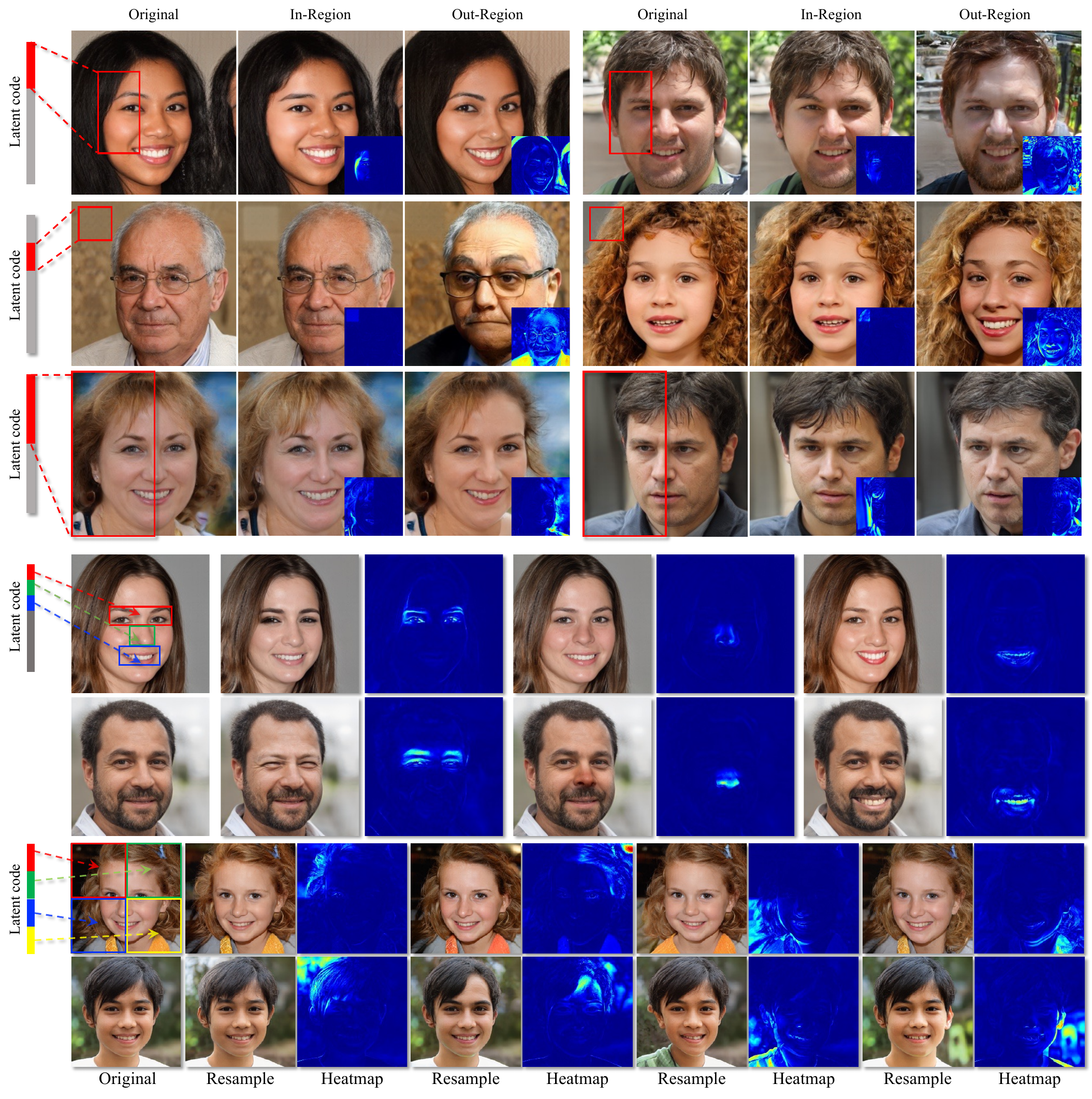}
  \vspace{-5pt}
  \caption{
    \textbf{Linking latents to some fixed regions} on human faces, which are pre-selected before training and shared by all instances.
    Linked latent subspaces and regions are highlighted with the same colors, the heatmaps reflect the change of pixel values after in-region resampling and out-region resampling.
    The results on the top group show linking a single region, while the results on the bottom show we can link multiple regions.
  }
  \label{fig:res-ffhq}
  \vspace{-10pt}
\end{figure*}

%%%% Figure: Results on the car.
\begin{figure*}[t]
  \centering
  \includegraphics[width=1.0\linewidth]{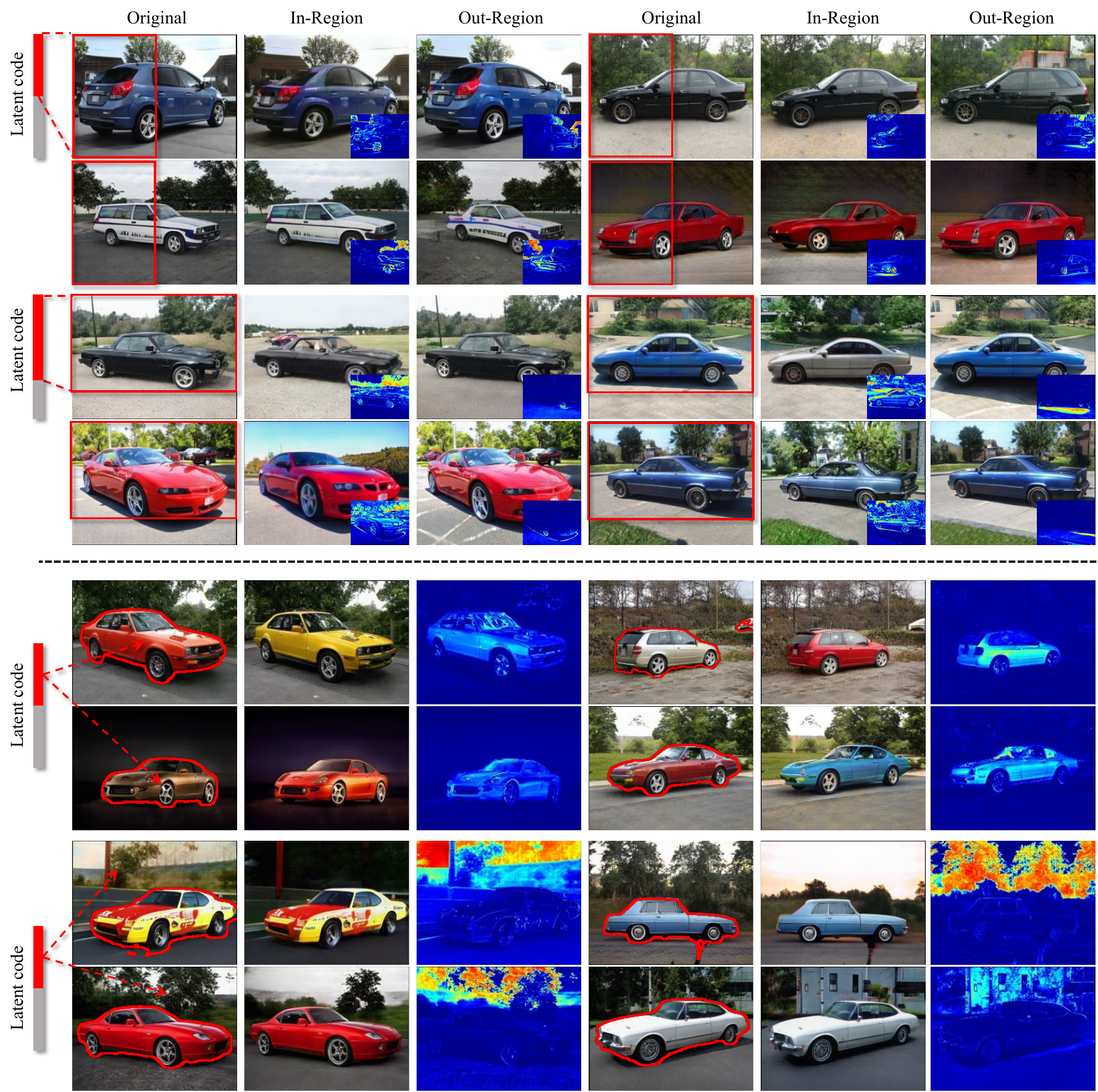}
  \vspace{-5pt}
  \caption{
    \textbf{Linking latents to regions} on cars.
    The regions in the top group are pre-selected before training and shared by all instances, while the regions in the bottom group dynamically vary across instances.
    Linked latent subspaces and regions are highlighted with \textcolor{red}{red} fragments and boxes/contours, the heatmaps reflect the change of pixel values after in-region resampling and out-region resampling.
    We can see that \method can link arbitrary regions no matter whether they are fixed or dynamically vary. 
  }
  \label{fig:res-car}
  \vspace{-10pt}
\end{figure*}

%%%% Figure: Results on the church.
\begin{figure*}[t]
  \centering
  \includegraphics[width=1.0\linewidth]{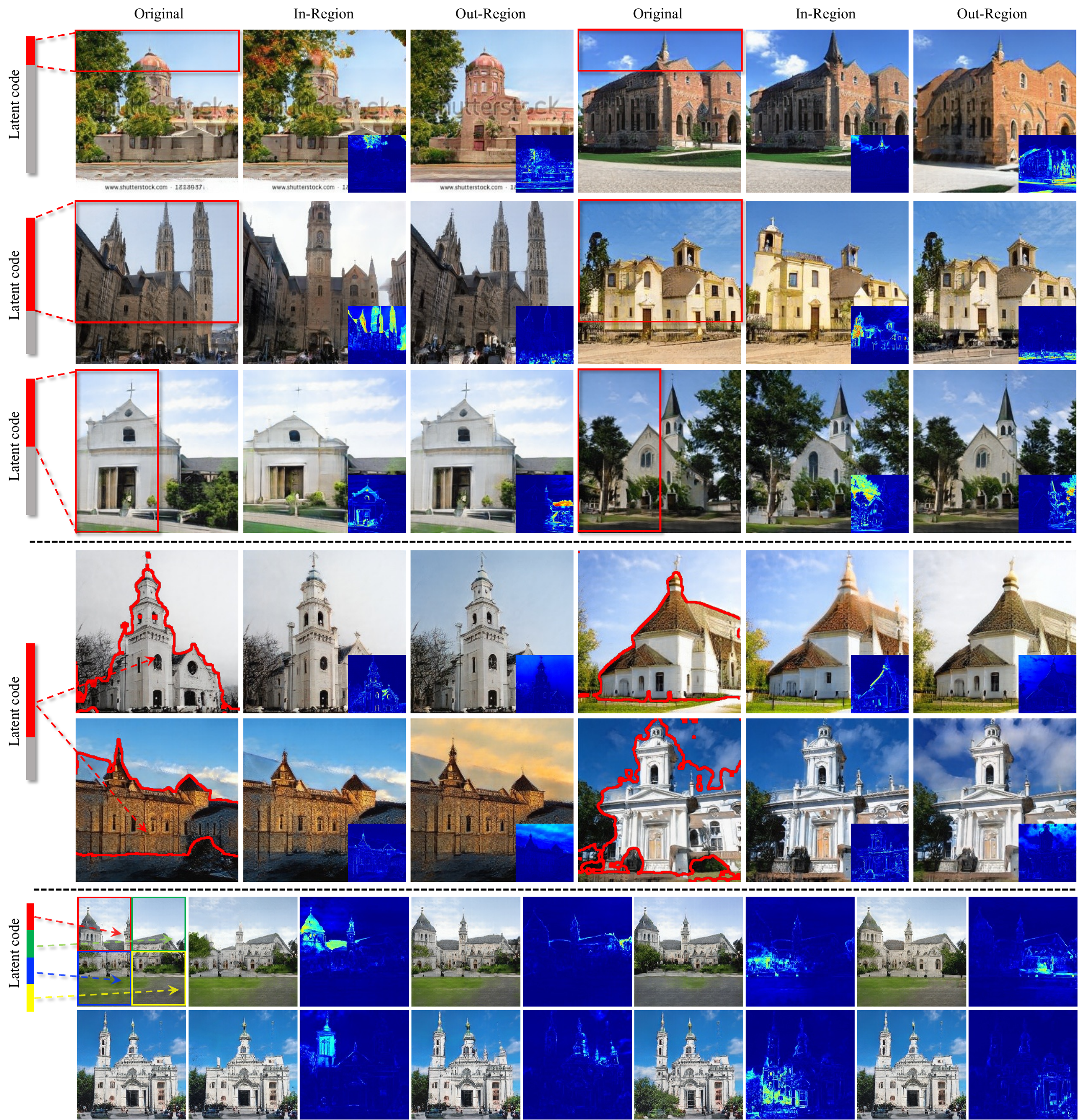}
  \vspace{-5pt}
  \caption{
    \textbf{Linking latents to regions} on church.
    The regions in the top group are pre-selected before training and shared by all instances, while the regions in the middle group dynamically vary across instances, and the bottom group shows we can link all the image regions to the latent space.
    Linked latent subspaces and regions are highlighted with different colors, the heatmaps reflect the change of pixel values after in-region resampling and out-region resampling.
    We can see that \method can link arbitrary regions no matter whether they are fixed, or dynamically vary, or even for the whole images.
  }
  \label{fig:res-church}
  \vspace{-5pt}
\end{figure*}

%%%% Figure: Results on the bedroom.
\begin{figure*}[t]
  \centering
  \includegraphics[width=0.90\linewidth]{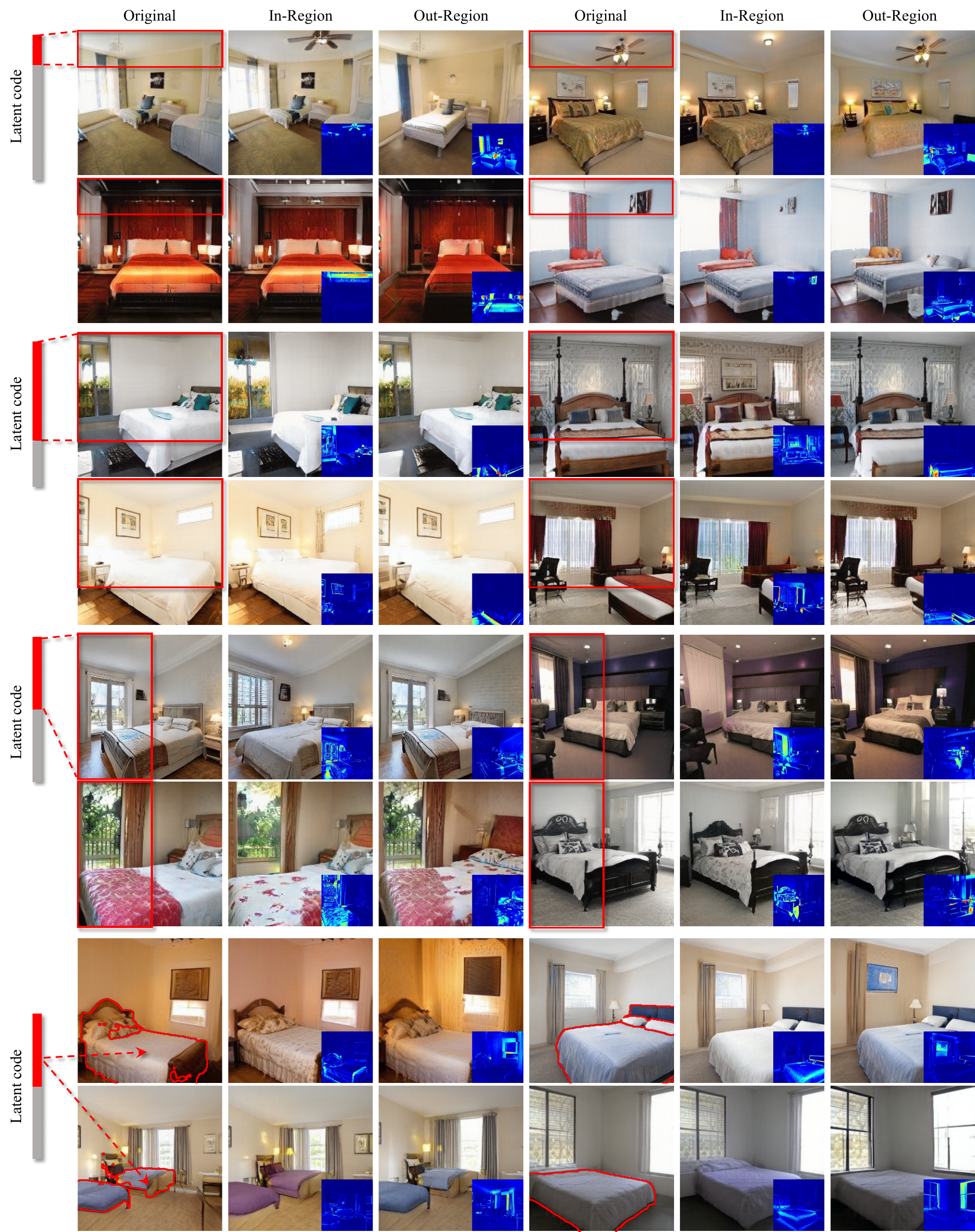}
  % \vspace{-5pt}
  \caption{
    \textbf{Linking latents to regions} on bedroom.
    The regions in the top group are pre-selected before training and shared by all instances, while the regions in the bottom group dynamically vary across instances.
    Linked latent subspaces and regions are highlighted with \textcolor{red}{red} fragments and boxes/contours, the heatmaps reflect the change of pixel values after in-region resampling and out-region resampling.
    We can see that \method can link arbitrary regions no matter whether they are fixed or dynamically vary. 
  }
  \label{fig:res-bedroom}
  \vspace{-10pt}
\end{figure*}

%%%% Figure: Results on EG3D.
\begin{figure*}[t]
  \centering
  \includegraphics[width=1\linewidth]{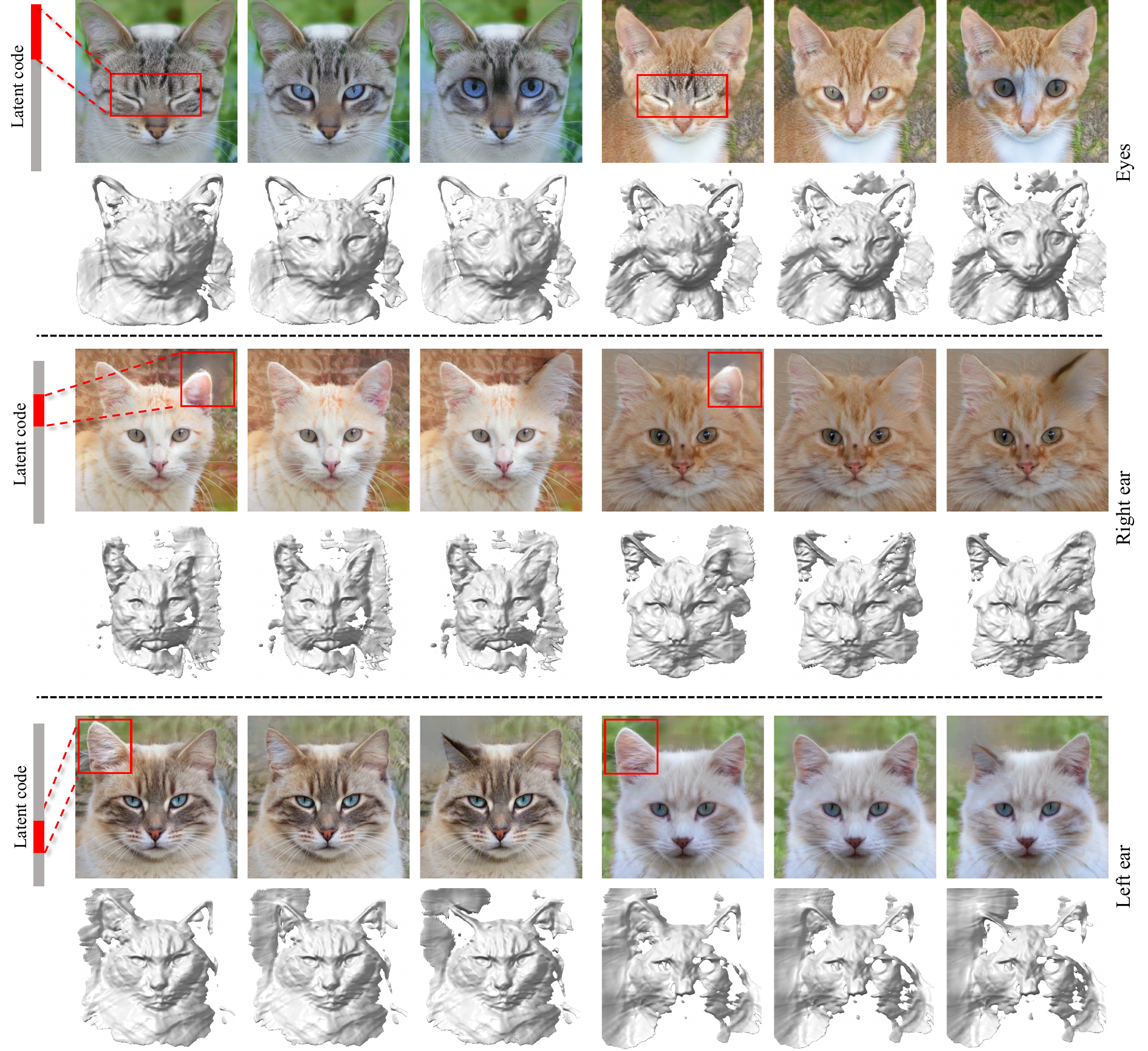}
  \caption{
    \textbf{Controllability on 3D-aware generative model}, \textit{i.e.}, EG3D~\cite{Chan2022eg3d}, under the cases of eyes,  left ear, and right ear.
    We find that \method is well compatible with 3D-aware image synthesis and allows controlling both the appearance and the underlying geometry.
  }
  \label{fig:res-eg3d}
\end{figure*}

\noindent\textbf{Comparison with existing methods.}
In our main text, we give the quantitative comparison results with some existing methods (Tab.2 in the main text).
Here, we show the corresponding qualitative results.
~\cref{fig:compare-eyes}, ~\cref{fig:compare-nose}, and~\cref{fig:compare-mouth} give the comparison results with ReSeFa~\cite{zhu2022resefa} and StyleCLIP~\cite{styleclip} on eyes, nose, and mouth regions. 
It is noteworthy that the ReSeFa is re-implemented on our fine-tuned model, and the text prompt we used in StyleCLIP when editing these three regions are: ``extremely big eyes without any change in the background'', ``crooked nose without any change in the background'', and ``open mouth without any change in the background''.
From these three figures, we can see that our method can achieve better control precision compared to the other two.
For instance, when editing the eyes, from the heatmaps, we can observe that the outlines of the face for ReSeFa and StyleCLIP are obviously changed.
The same phenomenon also occurs when editing the nose and mouth regions.
On the contrary, our method has negligible changes in the regions when editing a specific region thanks to our explicit link.
For Stylespace~\cite{wu2020stylespace}, we use the code released \href{https://github.com/betterze/StyleSpace/blob/main/StyleSpace_advance.ipynb/}{here} to manipulate eyes, nose, and mouth.
And the results are shown in~\cref{fig:compare-stylespace}, we can see from this figure, the outline of the face is also affected.
Hence, both quantitative (Tab.2 in the main text) and qualitative results demonstrate the precise control ability of our method.

\noindent\textbf{More results of our methods.}
We first give more quantitative results when our regularizer is added to different regions and datasets. 
~\cref{tab:fid-supp-single-region} and ~\cref{tab:fid-multi-region} report the FID on a single region and multiple regions, respectively.
From~\cref{tab:fid-supp-single-region}, we can see that on some datasets and regions, the FID slightly deteriorates (\textit{e.g.}, the regions on FFHQ), and on some datasets and regions, FID is comparable, even lower than without adding our regularizer (\textit{e.g.}, FFHQ on EG3D).
We believe that a higher FID is caused by the loss of diversity because our model requires the output image to be realistic before and after local editing (\textit{i.e.}, through partially resampling the latent code).
Such a hypothesis can be confirmed, to some degree, by the higher precision (\textit{i.e.}, image quality) and lower recall (\textit{i.e.}, diversity) shown in~\cref{tab:pr-supp}.
When more regions are linked, FID further deteriorates, as shown in~\cref{tab:fid-multi-region}.
It is sensible that with more constraints, more diversity will be lost.

~\cref{fig:res-ffhq}, ~\cref{fig:res-car}, ~\cref{fig:res-church}, ~\cref{fig:res-bedroom}, and ~\cref{fig:res-eg3d} show more qualitative results on FFHQ~\cite{stylegan}, AFHQ~\cite{choi2020starganv2}, LSUN-car, church, and bedroom~\cite{yu2015lsun}.
The linked regions in those figures are diverse and vary from a single fixed region to a dynamic change one, to multiple regions, and even to the whole images.
For example, on human faces, as shown in ~\cref{fig:res-ffhq}, we can link either complicated or non-special semantics (\textit{e.g.}, half of the faces, or just a cube of background) or link multiple regions to multiple latent fragments.
Results on the other datasets (\textit{e.g.}, \cref{fig:res-car}, ~\cref{fig:res-church}, ~\cref{fig:res-bedroom}) also demonstrate the success of our approach in linkage building regarding different regions or semantics.
Also, ~\cref{fig:res-eg3d} gives the results on the 3D generative model EG3D~\cite{Chan2022eg3d} trained on AFHQ~\cite{choi2020starganv2}, which is another evidence to demonstrate the generalization ability of our regularizer.

\setlength{\tabcolsep}{6.5pt}
\begin{table*}[t]
    \setlength{\tabcolsep}{4.5pt}
    \caption{
      Performance change after introducing our proposed regularizer into 2D and 3D baselines on a single region on different datasets, where the synthesis quality slightly drops but the controllability significantly improves.
    }
    \label{tab:fid-supp-single-region}
    \centering\small
    \begin{tabular}{l|cccccccccccccc|cc}
    \toprule
     Model &
      \multicolumn{13}{c|}{StyleGAN~\cite{stylegan2}} &
      \multicolumn{2}{c}{EG3D~\cite{Chan2022eg3d}}     \\ 
    \midrule
    Dataset &
      \multicolumn{3}{c|}{FFHQ} &
      \multicolumn{2}{c|}{AFHQ} &
      \multicolumn{3}{c|}{Church} &
      \multicolumn{3}{c|}{Car} &
      \multicolumn{2}{c|}{Bedroom} &
      \multicolumn{2}{c}{FFHQ} \\
    \midrule
    Region &
      \multicolumn{1}{c|}{Left} &
      \multicolumn{1}{c|}{Nose} &
      \multicolumn{1}{c|}{Mouth} &
      \multicolumn{1}{c|}{Eyes} &
      \multicolumn{1}{c|}{Ear} &
      \multicolumn{1}{c|}{Top} &
      \multicolumn{1}{c|}{Left} &
      \multicolumn{1}{c|}{Sky} &
      \multicolumn{1}{c|}{Bottom} &
      \multicolumn{1}{c|}{Left} &
      \multicolumn{1}{c|}{Car} &
      \multicolumn{1}{c|}{Top} &
      \multicolumn{1}{c|}{Bottom} &
      \multicolumn{1}{c|}{Nose} &
      \multicolumn{1}{c}{Mouth} \\ 
    \midrule
    \textit{w/o} Linking &
      \multicolumn{3}{c|}{3.98} &
      \multicolumn{2}{c|}{8.44} &
      \multicolumn{3}{c|}{3.82} &
      \multicolumn{3}{c|}{2.95} &
      \multicolumn{2}{c|}{3.01} &
      \multicolumn{2}{c}{4.28} \\ 
    \midrule
    LinkGAN (Ours) &
      \multicolumn{1}{l}{5.54} &
      \multicolumn{1}{l}{5.14} &
      \multicolumn{1}{l|}{5.11} &
      \multicolumn{1}{l}{10.52} &
      \multicolumn{1}{l|}{9.85} &
      \multicolumn{1}{l}{4.27} &
      \multicolumn{1}{l}{4.61} &
      \multicolumn{1}{l|}{4.40} &
      \multicolumn{1}{l}{2.88} &
      \multicolumn{1}{l}{3.07} &
      \multicolumn{1}{l|}{2.93} &
      \multicolumn{1}{l}{3.49} &
      \multicolumn{1}{l|}{3.72} &
      \multicolumn{1}{l}{4.17} & 4.21   \\ 
     \bottomrule
\end{tabular}
\end{table*}

\setlength{\tabcolsep}{6.5pt}
\begin{table}[t]
  \setlength{\tabcolsep}{4.5pt}
  \caption{
  Performance change after introducing our proposed regularizer on StyleGAN~\cite{stylegan2}  on multiple regions on different datasets.
  For FFHQ, we report links to two, three, and four regions, and for AFHQ and Church, we give the results of linking four regions (\textit{i.e.}, the whole image is linked).
  }
  \label{tab:fid-multi-region}
  \centering\small
  \begin{tabular}{l|ccc|c|c}
    \toprule
    Dataset
              & \multicolumn{3}{c|}{FFHQ} &  AFHQ  &  Church     \\
    \midrule
    \textit{w/o} Linking   
              & \multicolumn{3}{c|}{3.98}  &  8.44  &  3.82    \\
    \midrule
      LinkGAN  
              &  5.91   &  6.12   &  6.28  &  12.38  &  4.77    \\
    \bottomrule
  \end{tabular}
\end{table}

\setlength{\tabcolsep}{3.0pt}
\begin{table}[!ht]
    \centering\small
    \caption{
     Precision and Recall curve on different datasets trained on StyleGAN~\cite{stylegan2}.
    }
    \label{tab:pr-supp}
    \begin{tabular}{l|cc|cc|cc}
        \toprule
        & \multicolumn{2}{c|}{FFHQ}
        & \multicolumn{2}{c|}{AFHQ}
        & \multicolumn{2}{c}{Car}\\
        \midrule
        Metrics
        &  \Prec  & \Recall  &  \Prec  &  \Recall  &  \Prec  & \Recall  \\
        \midrule
        \textit{w/o} Linking
        &  0.849  &  0.200   &  0.847  &   0.126   &  0.826  &  0.294  \\
        LinkGAN
        &  0.864  &  0.165   &  0.844  &  0.021    &  0.838  &  0.252   \\
        \bottomrule
    \end{tabular}
\end{table}
\section{Ablation Study}\label{sec:abla}
\begin{figure}[t]
  \centering
  \includegraphics[width=1.0\linewidth]{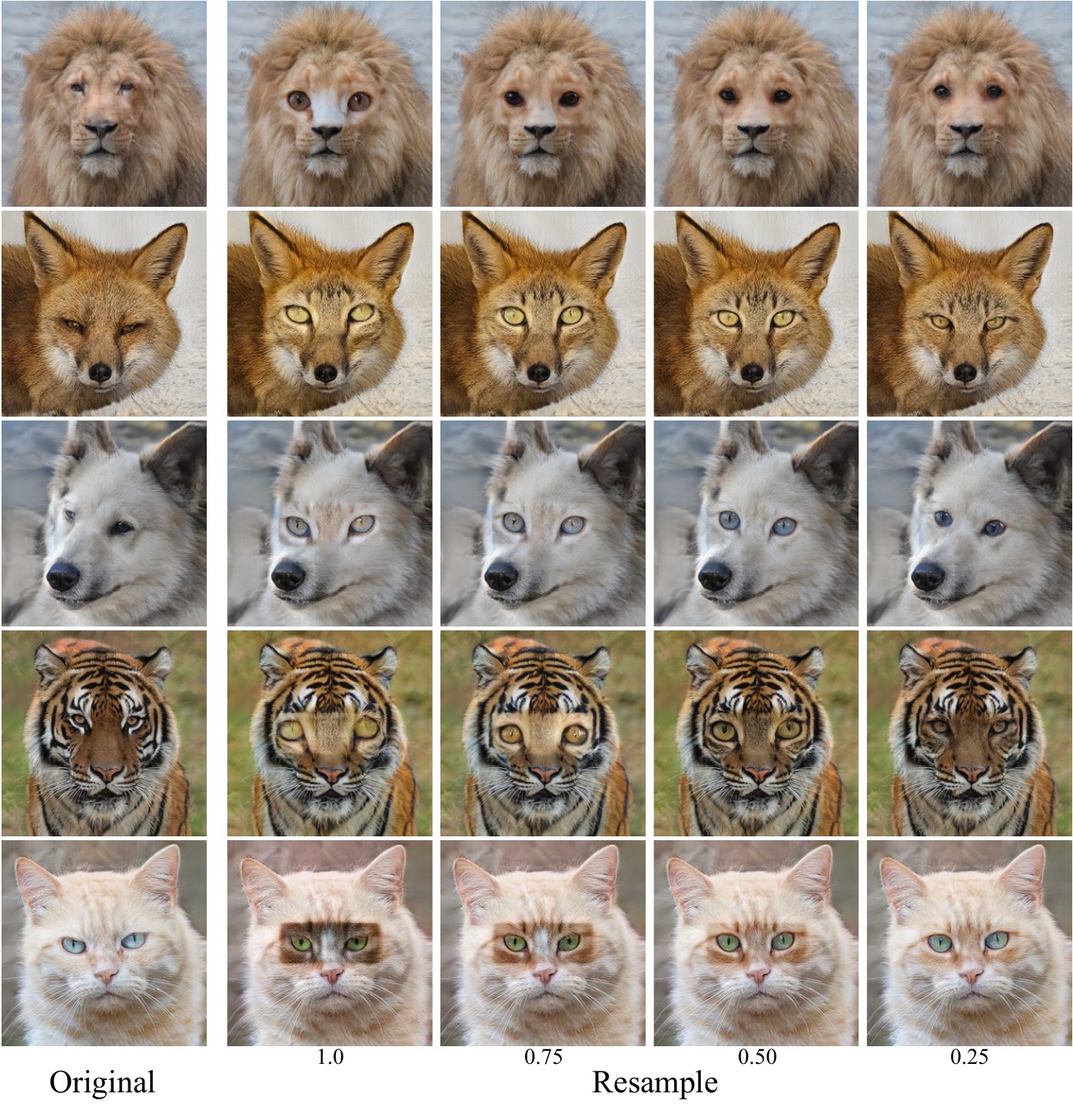}
  \vspace{-15pt}
  \caption{Ablation study on AFHQ~\cite{choi2020starganv2} using different interpolation strength. 
           The number under each column means the interpolation strength, \textit{e.g.}, 1 means the content in the eyes region is totally from resampled latent code. 
           In contrast, 0.25 means the content combines 0.75 original latent code and 0.25 of the resampled one.
           During training, the discriminator is not involved in the perturbed images.}
  \label{fig:ablation-interpolation-without-d}
  \vspace{-5pt}
\end{figure}

\begin{figure}[t]
  \centering
  \includegraphics[width=1.0\linewidth]{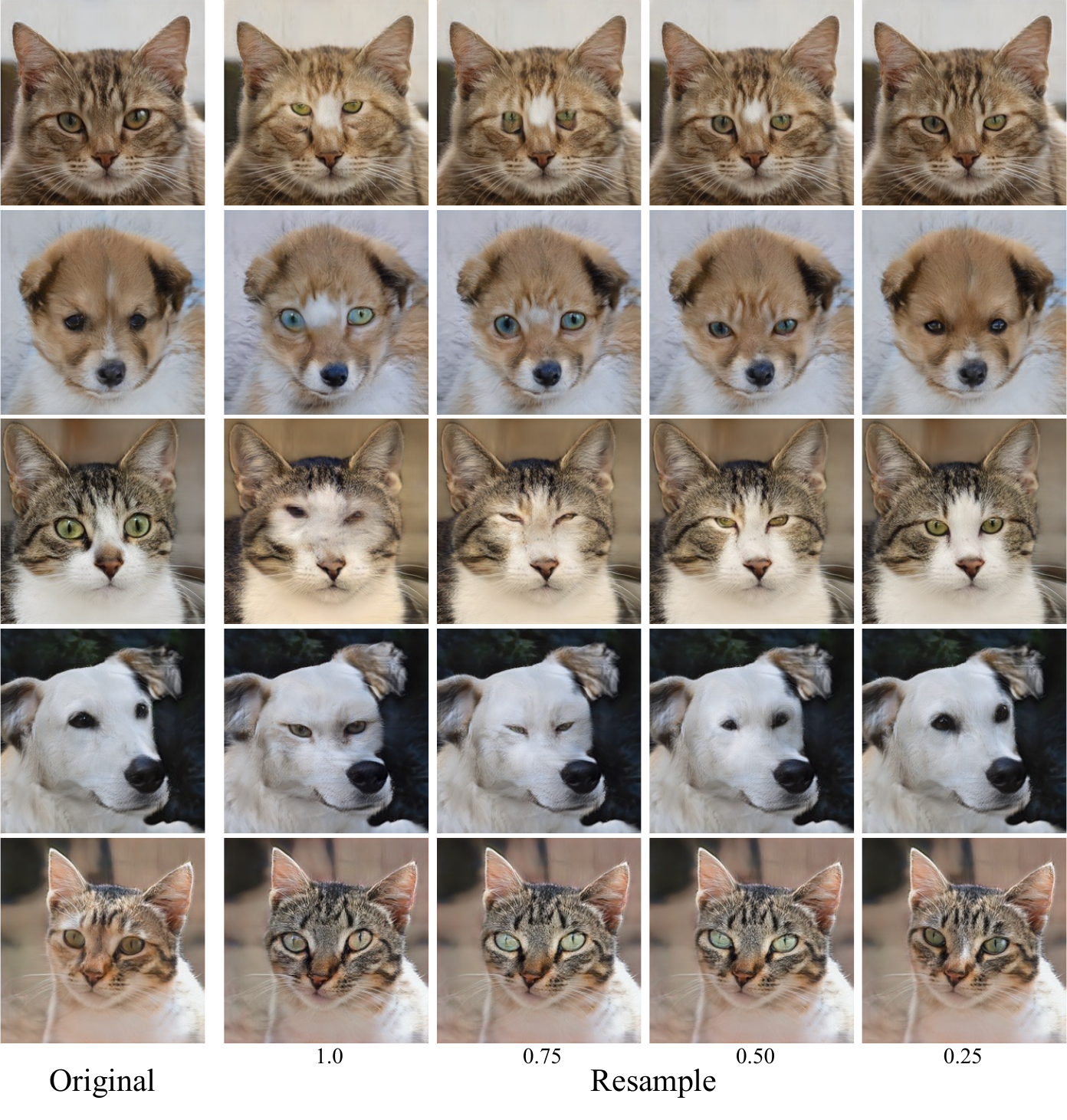}
  \vspace{-15pt}
  \caption{Ablation study on AFHQ~\cite{choi2020starganv2} using different interpolation strength. 
           The number under each column means the interpolation strength, \textit{e.g.}, 1 means the content in the eyes region is totally from resampled latent code. 
           In contrast, 0.25 means the content combines 0.75 original latent code and 0.25 of the resampled one.
           During training, the discriminator is involved in the perturbed images.}
  \label{fig:ablation-interpolation-with-d}
  \vspace{-5pt}
\end{figure}

Here we conduct an ablation study, which is the emergence of the inconsistency after resampling.
We do the study on the eye region of AFHQ~\cite{choi2020starganv2} dataset since it has multiple classes, and we can observe this inconsistency more clearly.
~\cref{fig:ablation-interpolation-without-d} and ~\cref{fig:ablation-interpolation-with-d} show some qualitative results when interpolating between the resampled part and the original latent part.
As we can see from these two figures, the inconsistency is obvious if we directly use a different latent code on the linked segments (\textit{i.e.}, the images with one on its bottom).
When we do the interpolating with the original latent code, the inconsistency can be alleviated, especially when we involve more than half of the content of the original latent code.
For instance, in the last row of~\cref{fig:ablation-interpolation-without-d}, when the color of the resampled cat is different from the original one, the resulting image shows severe inconsistency.
When we mix some content from the original latent code, the inconsistency is relieved.
Hence, interpolation is an effective way to remedy this inconsistency.

Another way to alleviate this inconsistency we explored is using the discriminator on the perturbed images.
Namely, when finetuning, we could involve the discriminator in those perturbed images to hinder the generator from synthesizing those inconsistency perturbed images.
We can see from the second column of~\cref{fig:ablation-interpolation-without-d} and~\cref{fig:ablation-interpolation-with-d}, which give a clear comparison.
For instance, in~\cref{fig:ablation-interpolation-without-d} we can see some edges of the rectangle in the eye region clearly, even the replaced content is aligned.
Instead, we can get much smoother resample results when involving the discriminator.
For example, the edge of the rectangle is disappeared, even when the resampled content is not well-aligned.
Hence, involving a discriminator is another effective way to alleviate inconsistency.

%%%% Figure

%%%% Figure

\section{Discussion and Conclusion}\label{sec:conclusion-supp}
After linking an arbitrary region to some latent axes with the size of $n$, any perturbation with randomly sampled $n$ dimension vector on the linked subspace results in the content change only in the linked image region, which can be viewed as a local semantic direction since it only influences the connected region.
However, some of the sampled latent vectors can not generate realistic manipulation, and some can (\textit{i.e.}, identical to the inconsistency phenomenon after resampling).
Hence, we need to verify whether the randomly sampled vector can produce a meaningful manipulation posteriorly, just like other unsupervised methods~\cite{ganspace, shen2021closed}.

\end{document}